\documentclass[journal]{IEEEtran}

\usepackage[utf8]{inputenc} 
\usepackage[T1]{fontenc}    
\usepackage{hyperref}       
\usepackage{url}            
\usepackage{booktabs}   
\usepackage{caption}
\usepackage[flushleft]{threeparttable}
\usepackage{subcaption} 
\usepackage{amsfonts}       
\usepackage{amsmath}
\usepackage{dsfont}
\usepackage{bbm}
\usepackage{amsthm}
\usepackage{nicefrac}       
\usepackage{microtype}      
\usepackage{lipsum}
\usepackage{algorithm}
\usepackage{mathtools}
\usepackage{algpseudocode}
\usepackage{graphicx} 
\usepackage{xcolor}
\graphicspath{{figures/}} 

\newcommand{\bs}{\boldsymbol}
\newcommand{\x}{\boldsymbol{x}}
\newcommand{\m}{\boldsymbol{m}}
\newcommand{\p}{\boldsymbol{p}}
\newcommand{\xib}{\bs{\xi}}
\newcommand{\thetab}{\boldsymbol{\theta}}
\newcommand{\phib}{\boldsymbol{\phi}}

\DeclareMathOperator*{\argmin}{argmin}
\newcommand{\given}{\,|\,}
\newcommand{\model}{\mathcal{M}_j}

\definecolor{RoyalBlue}{rgb}{0.1,0.45,0.75}

\title{Amortized Bayesian Model Comparison with Evidential Deep Learning}

\author{
  Stefan T.~Radev, Marco D'Alessandro, Ulf K.~Mertens, Andreas Voss, Ullrich Köthe, Paul-Christian Bürkner}

\begin{document}

\maketitle

\begin{abstract}
Comparing competing mathematical models of complex natural processes is a shared goal among many branches of science. The Bayesian probabilistic framework offers a principled way to perform model comparison and extract useful metrics for guiding decisions. 
However, many interesting models are intractable with standard Bayesian methods, as they lack a closed-form likelihood function or the likelihood is computationally too expensive to evaluate. With this work, we propose a novel method for performing Bayesian model comparison using specialized deep learning architectures. 
Our method is purely simulation-based and circumvents the step of explicitly fitting all
alternative models under consideration to each observed dataset. 
Moreover, it requires no hand-crafted summary statistics of the data and is designed to amortize the cost of simulation over multiple models, datasets, and dataset sizes.
This makes the method especially effective in scenarios where model fit needs to be assessed for a large number of datasets, so that case-based inference is practically infeasible.
Finally, we propose a novel way to measure epistemic uncertainty in model comparison
problems. We demonstrate the utility of our method on toy examples and simulated data from non-trivial models from cognitive science and single-cell neuroscience. We show that our method achieves excellent results in terms of accuracy, calibration, and efficiency across the examples considered in this work. We argue that our framework can enhance and enrich model-based analysis and inference in many fields dealing with computational models of natural processes. We further argue that the proposed measure of epistemic uncertainty provides a unique proxy to quantify absolute evidence even in a framework which assumes that the true data-generating model is within a finite set of candidate models.
\end{abstract}

\section{Introduction}

\noindent Researchers from various scientific fields face the problem of selecting the most plausible theory for an empirical phenomenon among multiple alternative theories. These theories are often formally stated as mathematical models which describe how observable quantities arise from unobservable (latent) parameters. Focusing on the level of mathematical models, the problem of theory selection then becomes one of \textit{model selection}.  

For instance, neuroscientists might be interested in comparing different models of spiking patterns given \textit{in vivo} recordings of neural activity \cite{izhikevich2004model}. 
Epidemiologists, on the other hand, might consider different models for predicting the spread and dynamics of an unfolding infectious disease \cite{verity2020estimates}. 
Crucially, the preference for one model over alternative models in these examples can have important consequences for research projects or social policies. 

Accounting for complex natural phenomena often requires specifying complex models which entail some degree of randomness. 
Inherent stochasticity, incomplete  description, or epistemic ignorance all call for some form of uncertainty awareness. 
To make matters worse, empirical data on which models are fit are necessarily finite and can only be acquired with finite precision. Finally, the plausibility of many non-trivial models throughout various branches of science can be assessed only approximately, through expensive simulation-based methods \cite{ratmann2009model, deisboeck2010multiscale, turner2016bayesian, marin2018likelihood, izhikevich2004model, da2018model}. 

Ideally, a method for approximate model comparison should meet the following desiderata:
\begin{enumerate}
    \item \emph{Theoretical guarantee}: Model probability estimates should be, at least in theory, calibrated to the true model probabilities induced by an empirical problem;
    \item \emph{Accurate approximation}: Model probability estimates should be accurate even for finite or small sample sizes;
    \item \emph{Occam's razor}: Preference for simpler models should be expressed by the model probability estimates;
    \item \emph{Scalability}: The method should be applicable to complex models with implicit likelihood within reasonable time limits;
    \item \emph{Efficiency}: The method should enable fully amortized inference- over arbitrarily many models, datasets and different dataset sizes;
    \item \emph{Maximum data utilization}: The method should capitalize on all information contained in the data and avoid information loss through insufficient summary statistics of the data;
\end{enumerate}
In the current work, we address these desiderata with a novel method for Bayesian model comparison based on evidential deep neural networks. 
Our method works in a purely simulation-based manner and circumvents the step of separately fitting all alternative models to each dataset. 
To this end, for any particular model comparison problem, we propose to train a \textit{specialized expert network} which encodes global information about the generative scope of each model family. 
In this way, Bayesian model comparison is amortized over multiple models, datasets, and dataset sizes, which makes our method applicable in scenarios where case-based inference is way too costly to perform with standard methods (cf. Figure \ref{fig:framework}).

In addition, we propose to avoid hand-crafted summary statistics (a feature on which standard methods for simulation-based inference heavily rely) by utilizing novel deep learning architectures which are aligned to the probabilistic structure of the raw data (e.g., permutation invariant networks \cite{bloem2019probabilistic}, recurrent networks \cite{gers1999learning}). 

Finally, we explore a novel way to measure epistemic uncertainty in model comparison problems, following the pioneering work of \cite{sensoy2018evidential} on image classification.
We argue that this measure of epistemic uncertainty provides a unique proxy to quantify absolute evidence even in an $\mathcal{M}$-closed framework, which assumes that the true data-generating model is within the candidate set \cite{yao2018using}.

\section{Background}

\subsection{Bayesian Inference}

\noindent A consistent mathematical framework for describing uncertainty and quantifying model plausibility is offered by the Bayesian view on probability theory \cite{jaynes2003probability}. 
In a Bayesian setting, we start with a collection of $J$ competing generative models $\mathcal{M} = \{\mathcal{M}_1,\mathcal{M}_2,\dots, \mathcal{M}_J\}$. 
Each $\mathcal{M}_j$ is associated with a generative mechanism $g_j$, typically realized as a Monte Carlo simulation program, and a corresponding parameter space $\Theta_j$.
Ideally, each $g_j$ represents a theoretically plausible (potentially noisy) mechanism by which observable quantities $\x$ arise from hidden parameters $\thetab$ and independent noise $\bs{\xi}$:
\begin{align}
     \x = g_{j}(\thetab_j, \bs{\xi}) \textrm{ with } \thetab_j \in \Theta_{j} \label{eq:1}
\end{align}
where $\Theta_{j}$ is the corresponding parameter space of model $g_{j}$ and the subscript $j$ explicates that each model might be specified over a different parameter space. 
We assume that the functional or algorithmic form of each $g_{j}$ is known and that we have a sample (dataset) $\{\x_i\}_{i=1}^N := \x_{1:N}$ of $N$ (multivariate) observations $\x_n \in \mathcal{X}$ generated from an unknown process $p^*$. 
The task of Bayesian model selection is to choose the model in $\mathcal{M}$ that best describes the observed data by balancing simplicity (sparsity) and predictive performance.

\subsection{The Likelihood}

\noindent A central object in Bayesian inference is the \textit{likelihood function}, denoted as $p(\x \given \thetab_j, \mathcal{M}_j)$.
Loosely speaking, the likelihood returns the relative probability of an observation observation $\x$ (or a sequence of observations $\x_{1:N}$) given a parameter configuration $\thetab_j$ and model assumptions $\mathcal{M}_j$. 
When the parameters are systematically varied and the data held constant, the likelihood can be used to quantify how well each model instantiation fits the data. 

If the likelihood of a generative model can be associated with a known probability density function (e.g., Gaussian), the model can be formulated entirely in terms of the likelihood and the likelihood can be evaluated analytically or numerically for any pair $(\x, \thetab)$. 
On the other hand, if the likelihood is unknown or intractable, as is the case when dealing with complex models, one can still generate random samples from the model by running the simulation program with a random configuration of its parameters. 

This is due to the fact that each stochastic model, viewed as a Monte Carlo simulator, defines an implicit likelihood given by the relationship: 
\begin{equation}
    \p(\x \given \thetab_j, \mathcal{M}_j) = \int_{\Xi} \delta(\x - g_j(\thetab_j, \bs{\xi}))\,p(\bs{\xi} \given \thetab_j)\,d \bs{\xi}
\end{equation}
where $\delta(\cdot)$ is the Dirac delta function and the integral runs over all possible execution paths of the stochastic simulation for a fixed $\thetab_j$.
For most complex models, this integral is analytically intractable or too expensive to approximate numerically, so it is much easier to specify the model directly in terms of the simulation program $g_j$ instead of deriving the likelihood $p(\x \given \thetab_j, \mathcal{M}_j)$. 
Importantly, we can still \textit{sample} from the likelihood by running the simulator with different Monte Carlo realizations of $\bs{\xi}$, that is, for a fixed $\theta_j$, we have the following equivalence:
\begin{align}
     \x_n \sim p(\x \given \thetab_j,\mathcal{M}_j) \Longleftrightarrow\ \x_n = g_{j}(\thetab_j,\bs{\xi}_n) \textrm{ with } \bs{\xi}_n \sim p(\bs{\xi}) \label{eq:2}
\end{align}

\subsection{Bayes Factors}

\noindent How does one assign preferences to competing models using a Bayesian toolkit? 
The canonical measure of evidence for a given model is the \textit{marginal likelihood}:
\begin{equation}
    p(\x_{1:N} \given \mathcal{M}_j) = \int_{\Theta_j}p(\x_{1:N} \given \thetab_j, \mathcal{M}_j)\,p(\thetab_j \given \mathcal{M}_j)\,d\thetab_j \label{eq:marg_lik}
\end{equation}
which is, in general, intractable to compute for non-trivial models.
Importantly, the dependence on the prior over model $\mathcal{M}_j$'s parameters introduces a probabilistic version of Occam's razor, which expresses our preference for a simpler model over a more complex one when both models can account for the data equally well.
The marginal likelihood thus focuses on \textit{prior predictions} and penalizes the prior complexity of a model (i.e., the prior acts as a weight on the likelihood).
This is in contrast to \textit{posterior predictions}, which require marginalization over the parameter posterior $p(\thetab_j \given \x_{1:N}, \mathcal{M}_j)$ and can be used to select the model which best predicts new data.

Provided that the marginal likelihood can be efficiently approximated, one can compute the ratio of marginal likelihoods for two models $\mathcal{M}_j$ and $\mathcal{M}_k$ via
\begin{equation}
    \textrm{BF}_{jk} = \frac{p(\x_{1:N} \given \mathcal{M}_j)}{p(\x_{1:N} \given \mathcal{M}_k)}. \label{eq:bf}
\end{equation}
This famous ratio is called a Bayes factor (BF) and is used in Bayesian settings for quantifying relative model preference.
Thus, a $\textrm{BF}_{jk}$ $> 1$ indicates preference for model $j$ over model $k$, given a set of observations $\x_{1:N}$.
Alternatively, one can directly focus on the (marginal) posterior probability of a model $\mathcal{M}_j$,
\begin{equation}
    p(\mathcal{M}_j \given \x_{1:N}) \propto p(\x_{1:N} \given \mathcal{M}_j)\,p(\mathcal{M}_j) \label{eq:postodds}
\end{equation}
which equips the model space itself with a prior distribution $p(\mathcal{M})$ over the considered model space encoding potential preferences for certain models before collecting any data.
Such a prior might be useful if a model embodies extraordinary claims (e.g., telekinesis) and thus requires extraordinary evidence supporting it. 
However, if no prior reasons can be given for favoring some models over others (i.e., one prefers not to prefer), a uniform model prior $p(\mathcal{M}) = 1/J$ can be assumed.

The ratio of posterior model probabilities is called the \textit{posterior odds} and is connected to the Bayes factor via the corresponding model priors:
\begin{equation}
    \frac{p(\mathcal{M}_j \given \x_{1:N})}{p(\mathcal{M}_k \given \x_{1:N})} = \frac{p(\x_{1:N} \given \mathcal{M}_j)}{p(\x_{1:N} \given \mathcal{M}_k)} \times \frac{p(\mathcal{M}_j)}{p(\mathcal{M}_k)}
\end{equation}
If two models are equally likely \textit{a priori}, the posterior odds equal the Bayes factor. 
In this case, if the Bayes factor, or, equivalently, the posterior odds equal one, the observed data provide no decisive evidence for one of the models over the other. 
However, a relative evidence of one does not allow to distinguish whether the data are equally likely or equally unlikely under both models, as this is a question of absolute evidence.
Needless to say, the distinction between relative and absolute evidence is of paramount importance for model comparison, so we address it in the next section on model comparison frameworks.

\subsection{$\mathcal{M}$-Frameworks}

\noindent In Bayesian inference, the relationship between the true generative process $p^*$ and the model list $\mathcal{M}$ can be classified into three categories: $\mathcal{M}$-closed, $\mathcal{M}$-complete and $\mathcal{M}$-open \cite{yao2018using}. 
Closely related to the distinction between relative and absolute evidence is the distinction between $\mathcal{M}$-closed and $\mathcal{M}$-complete frameworks. 
Under an $\mathcal{M}$-closed framework, the true model is assumed to be in the predefined set of competing models $\mathcal{M}$, so relative evidence \textit{is} identical to absolute evidence. Under an $\mathcal{M}$-complete framework, a true model is assumed to exist but is not necessarily assumed to be a member of $\mathcal{M}$. However, one still focuses on the models in $\mathcal{M}$ due to computational or conceptual limitations\footnote{In this work, we delegate the discussion of whether the concept of a true model has any ontological meaning to philosophy. See also \cite{yao2018using} for discussion of an $\mathcal{M}$-open framework, in which no true model is assumed to exist.}. 

Deciding on the particular $\mathcal{M}$-framework under which a model comparison problem is tackled is often a matter of prior theoretical considerations. 
However, since in most non-trivial research scenarios $\mathcal{M}$ is a finite set and candidate models in $\mathcal{M}$ are often simpler approximations to the true model, there will be \textit{uncertainty} as to whether the observed data could have been generated by one of these models. 
In the following, we will refer to this uncertainty as \textit{epistemic uncertainty}. 
Our method utilizes a data-driven way to calibrate its epistemic uncertainty in addition to the model probabilities through simulations under an $\mathcal{M}$-closed framework.

Consequently, given real observed data, a researcher can obtain a measure of uncertainty with regard to whether the generative model of the data is likely to be in $\mathcal{M}$ or not. From this perspective, our method lies somewhere between an $\mathcal{M}$-closed and an $\mathcal{M}$-complete framework as it provides information from both viewpoints. 
In this way, our approach to model misspecification differs from \textit{likelihood-tempering} methods, which require an explicit evaluation of a \textit{tilted} likelihood (raised to a power $0 < t < 1)$ in order to prevent overconfident Bayesian updates \cite{grunwald2017inconsistency}. 

\section{Related Work}

\noindent Bayesian methods for model comparison can be categorized as either posterior predictive or prior predictive approaches \cite{BDA3}, with our method falling into the latter category.
Posterior predictive approaches are concerned with predicting new data using models trained on the current data. 
In prior predictive approaches, models are conditioned only on prior information but not on the current data. 
Accordingly, all current data counts as new data for the purpose of prior predictive methods.

Naturally, cross-validation (CV) procedures are the main approach for posterior predictive comparisons \cite{vehtari2012}. 
Examples for widely applied methods that fall into this category are approximate cross-validation procedures using Pareto-smoothed importance sampling \cite{vehtari2017practical, burkner2020LFO}, information criterion approaches such as the widely applicable information criterion (WAIC; \cite{watanabe2013waic}), or stacking of posterior predictive distributions \cite{yao2018using}.

All of these methods require not only the ability to evaluate the likelihood of each model for each observation during parameter estimation, but also for new observations during prediction. 
What is more, if application of exact CV methods is required because approximations are insufficient or unavailable, models need to be estimated several times based on different data sets or subsets of the original data set.
This renders such methods practically infeasible when working with complex simulators for which estimating models even once is already very slow. 
Thus, even a single intractable model in the model set suffices to disproportionately increase the difficultly of performing model comparison.


In contrast, our proposed method circumvents explicit parameter estimation and focuses directly on the efficient approximation of Bayes factors (or posterior model probabilities). 
Moreover, it overcomes two major sources of intractability that stand in the way of Bayesian model comparison via Bayes factors: the likelihood (Eq.\ref{eq:2}) and the marginal likelihood (Eq.\ref{eq:marg_lik}).

When the likelihood can be computed in closed-form, sophisticated algorithms for efficiently approximating the (intractable) marginal likelihood have been proposed in the Bayesian universe, such as \textit{bridge sampling} and \textit{path sampling} \cite{gelman1998simulating, gronau2017tutorial}.
However, these methods still depend on the ability to evaluate the likelihood $p(\x \given \thetab_j, \mathcal{M}_j)$ for each candidate model. 
If, in addition, the likelihood itself is intractable, as is the case with complex simulators, researchers need to resort to expensive simulation-based methods \cite{turner2014generalized, turner2016bayesian, marin2018likelihood, papamakarios2018sequential}. 

A standard set of tools for Bayesian simulation-based inference is offered by approximate Bayesian computation (ABC) methods \cite{sunnaaker2013approximate, mertens2018abrox}. 
ABC methods approximate the model posterior by repeatedly sampling parameters from each proposal (prior) distribution and then simulating multiple datasets by running each simulator with the sampled parameters.
A pre-defined similarity criterion determines whether a simulated dataset (or a summary statistic thereof) is sufficiently similar to the actually observed dataset. 
The model that most frequently generates synthetic observations matching those in the observed dataset is the one favored by ABC model comparison. 

Despite being simple and elegant, standard ABC methods involve a crucial trade-off between accuracy and efficiency. 
In other words, stricter similarity criteria yield more accurate approximations of the desired posteriors at the price of higher and oftentimes intolerable rejection rates. 
What is more, most ABC methods require multiple \textit{ad hoc} decisions from the method designer, such as the choice of similarity criterion or the summary statistics of the data (e.g., moments of empirical distributions) \cite{marin2018likelihood}. 
However, there is no guarantee that hand-crafted summaries extract all relevant information and model comparison with insufficient summary statistics can dramatically deteriorate the resulting model posteriors \cite{robert2011lack}. 
More scalable developments from the ABC family (ABC-SMC, ABC-MCMC, ABC neural networks and the recently proposed ABC random forests) offer great efficiency boosts but still rely on hand-crafted summary statistics \cite{marin2018likelihood, jiang2017learning, sisson2011likelihood}.

Recently, a number of promising innovations from the machine learning and deep learning literature have entered the field of simulation-based inference \cite{cranmer2020frontier}. 
For instance, the sequential neural likelihood (SNL, \cite{papamakarios2018sequential}), the automatic posterior transformation (APT, \cite{greenberg2019automatic}), the amortized ratio estimation \cite{hermans2020likelihood} or the BayesFlow method \cite{radev2020bayesflow} all implement powerful neural density estimators to overcome the shortcomings of standard ABC methods.
Moreover, these methods involve some degree of \textit{amortization}, which ensures extremely efficient inference after a potentially costly upfront training phase. 
However, neural density estimation focuses solely on efficient Bayesian parameter estimation instead of scaling up Bayesian model comparison.
With certain caveats, neural density estimators can be adapted for Bayesian model comparison by post-processing the samples from an approximate posterior/likelihood over each model's parameters.
However, such an approach will involve training a separate neural estimator for each model in the candidate set and has not yet been systematically investigated.
In addition, most of these methods also rely on fixed summary statistics \cite{papamakarios2018sequential}) and few applications using raw data directly exist \cite{radev2020bayesflow, greenberg2019automatic}. 

\begin{figure*}
\centering
\begin{subfigure}{.99\textwidth}
    \includegraphics[width=\textwidth]{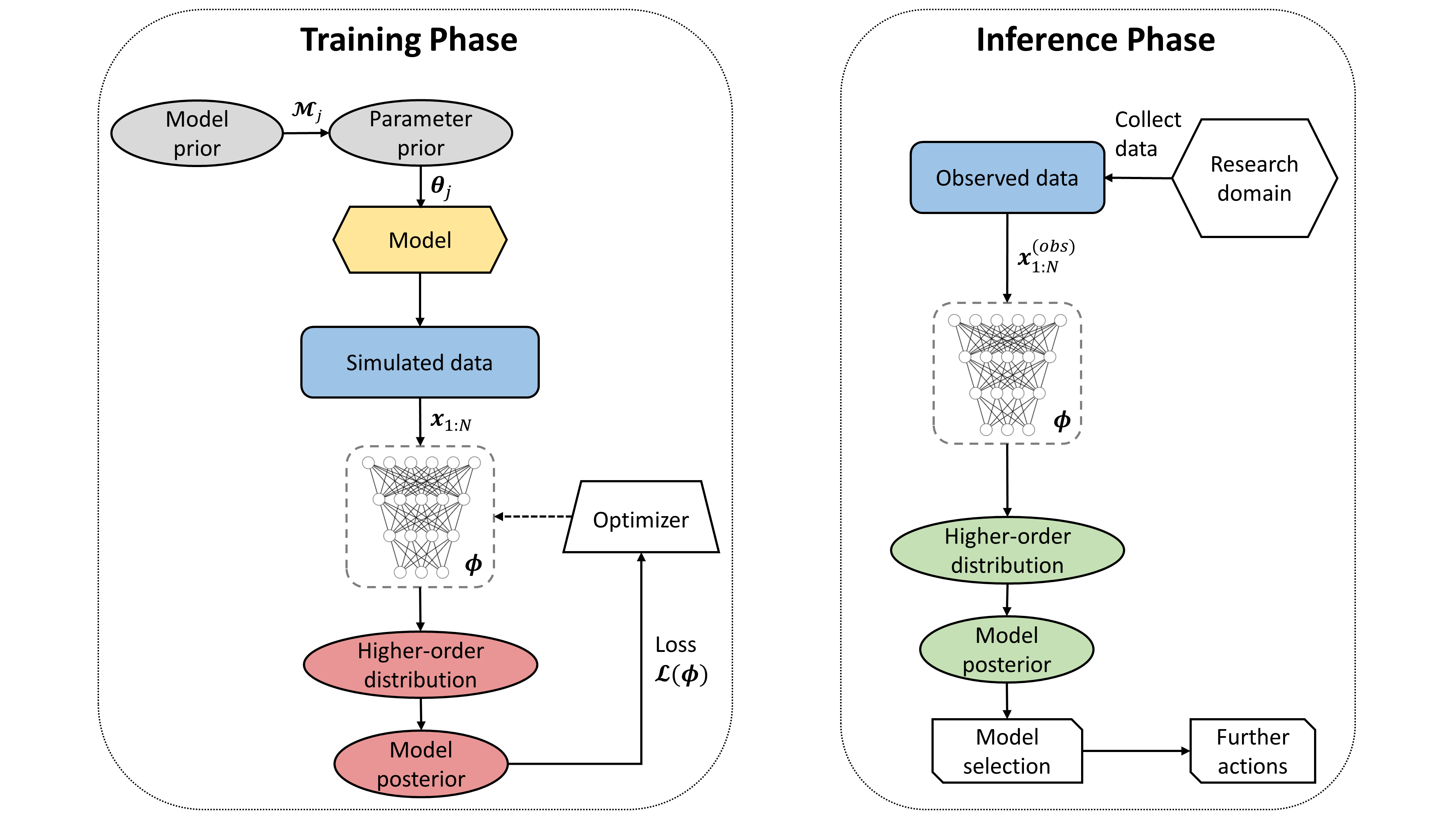}
\end{subfigure}
\caption[short]{Left panel: The simulation-based training phase of our evidential method. Right panel: The inference phase with real data and a pre-trained evidential network.} \label{fig:framework}
\end{figure*}

Alongside advancements in simulation-based inference, there has been an upsurge in the development of methods for uncertainty quantification in deep learning applications.
For instance, much work has been done on the efficient estimation of Bayesian neural networks \cite{hernandez2015probabilistic, li2017dropout, louizos2017multiplicative} since the pioneering work of \cite{mackay1995bayesian}.
Parallel to the establishment of novel variational methods \cite{kingma2019introduction, kingma2015variational}, these ideas have paved the way towards more interpretable and trustworthy neural network inference. 
Moreover, the need for distinguishing between different sources of uncertainty and the overconfidence of deep neural networks in classification and regression tasks has been demonstrated quite effectively \cite{kendall2017uncertainties, sensoy2018evidential}.
Our current work draws on recent methods for evidence and uncertainty representation in classification tasks \cite{sensoy2018evidential}. However, our goal is to efficiently approximate Bayes factors between competing mechanistic models using non-Bayesian neural networks, not to estimate neural network parameters (e.g., weights) via Bayesian methods.

Our method combines latest ideas from simulation-based inference and uncertainty quantification for building efficient and uncertainty-aware estimators for amortized Bayesian model comparison.
As such, it is intended to complement the toolbox of simulation-based methods for parameter estimation with crucial model comparison capabilities and incorporates some unique features beyond the scope of standard ABC methods.  
In the following, we describe the building blocks of our method.

\section{Evidential Networks for Bayesian Model Comparison}
\label{evidential-networks}

\subsection{Model Selection as Classification}

\noindent In line with previous simulation-based approaches to model selection, we will utilize the fact that we can generate arbitrary amounts of data via Eq.\ref{eq:2} for each considered model $\mathcal{M}_j$. Following \cite{pudlo2015reliable, marin2018likelihood}, we cast the problem of model comparison as a probabilistic classification task. In other words, we seek a parametric mapping $f_{\phib} : \mathcal{X}^N \rightarrow \Delta^J$ from an arbitrary data space $\mathcal{X}^N$ to a probability simplex $\Delta^J$ containing the posterior model probabilities $p(\mathcal{M} \given \x_{1:N})$.
Previously, different learning algorithms (e.g., random forests \cite{marin2018likelihood}) have been employed to tackle model comparison as classification. Following recent developments in algorithmic alignment and probabilistic symmetry \cite{xu2019what, bloem2019probabilistic}, our method parameterizes $f_{\bs{\phi}}$ via a specialized neural network with trainable parameters $\phib$ which is aligned to the probabilistic structure of the observed data. See the \textbf{Network Architectures} section in \textbf{Appendix A} for a detailed description of the employed networks' structure.

In addition, our method differs from previous classification approaches to model comparison in the following aspects. 
First, it requires no hand-crafted summary statistics, since the most informative summary statistics are learned directly from data. 
Second, it uses online learning (i.e., on-the-fly simulations) and requires no storage of large reference tables or data grids. Third, the addition of new competing models does not require changing the architecture or re-training the network from scratch, since the underlying data domain remains the same. 
In line with the transfer learning literature, only the last layer of a pre-trained network needs to be changed and training can be resumed from where it had stopped.
Last, our method is uncertainty-aware, as it returns a higher-order distribution over posterior model probabilities. From this distribution, one can extract both absolute and relative evidences, as well as quantify the model selection uncertainty implied by the observed data. 

To set up the model classification task, we run \hyperref[alg:1]{\textbf{Algorithm 1}} repeatedly to construct training batches with $B$ simulated data sets of size $N$ and $B$ model indices of the form $\mathcal{D}^{(B)}_N := \{(\m^{(b)}, \x_{1:N}^{(b)})\}_{b=1}^{B}$. 
We then feed each batch to a neural network which takes as input simulated data with variable sizes and returns a distribution over posterior model probabilities. 
The neural network parameters are optimized via standard backpropagation. 
Upon convergence, we can apply the pre-trained network to arbitrarily many datasets of the form $\x_{1:N}^{(obs)}$ to obtain a vector of probabilities $p_{\bs{\phi}}(\m \given \x_{1:N}^{(obs)})$ which approximates the true model posterior $p(\mathcal{M} \given \x_{1:N}^{(obs)})$.

\begin{algorithm*}
\caption{Monte Carlo generation of synthetic data sets for model comparison}\label{alg:1}
\begin{algorithmic}[1]
\Require {$p(\mathcal{M})$ - prior over models, $\{p(\thetab \given \mathcal{M}_j)\}$ - list of priors over model parameters, $\{g_j\}$ - list of stochastic simulators, $\{p_j(\xib)\}$ - list of noise distributions (RNGs), $p(N)$ - distribution over data set sizes, $B$ - number of data sets to generate (batch size)} 
\State {Draw data set size: $N \sim p(N)$}
\For{$b = 1,...,B$}
\State {Draw model index from model prior: $\model^{(b)} \sim p(\mathcal{M})$}
\State{Draw model parameters from prior: $\thetab_j^{(b)} \sim p(\thetab_j \given \model^{(b)})$}
\For{$n = 1,...,N$}
\State{Sample noise instance: $\xib_n \sim p_j(\xib)$}
\State{Run simulator $j$ to obtain $n$-th synthetic observation: $\x_n = g_j(\thetab_j^{(b)}, \xib_n)$}
\EndFor
\State{Encode model index as a one-hot-encoded vector: $\m^{(b)} = \textrm{OneHotEncode}(\model^{(b)})$}
\State{Store pair $(\m^{(b)}, \x^{(b)}_{1:N})$ in $\mathcal{D}_N^{(B)}$}
\EndFor
\State{\textbf{return} mini-batch $\mathcal{D}_N^{(B)} := \{\m^{(b)}, \x_{1:N}^{(b)})\}_{b=1}^{B}$}
\end{algorithmic}
\end{algorithm*}
Note, that this procedure incurs no memory overhead, as the training batches need not be stored in memory all at once. Intuitively, the connection between data and models is encoded in the network's weights. Once trained, the evidential network can be reused to perform instant model selection on multiple real observations. As mentioned above, the addition of new models requires simply adjusting the pre-trained network, which requires much less time than re-training the network from scratch. We now describe how model probabilities and evidences are represented by the evidential network. 

\begin{figure*}
\centering
\begin{subfigure}{.99\textwidth}
    \includegraphics[width=\textwidth]{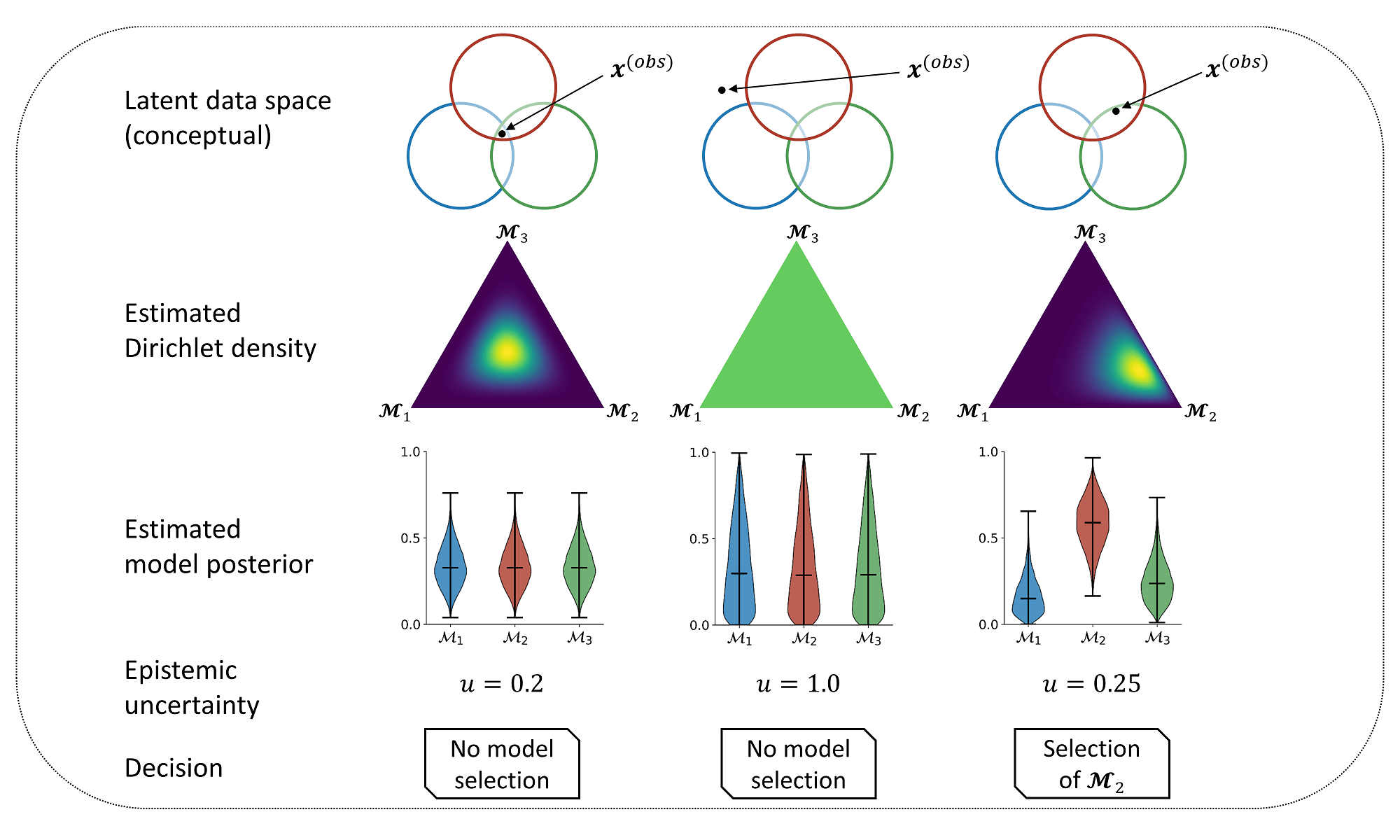}
\end{subfigure}
\caption[short]{Three different hypothetical model comparison scenarios with different observations. The first column depicts observing a dataset which is equally probable under all models. In this case, the best candidate model cannot be selected and the Dirichlet density peaks in the middle of the simplex. The second column depicts a dataset which is beyond the generative scope of all models and no model selection decision is possible. The Dirichlet density in this case is flat which indicates total uncertainty. The third column illustrates an observed dataset which is most probable under model 2, so the Dirichlet simplex is peaked towards the corner encoding model 2, and the corresponding model posterior for model 2 is highest.} \label{fig:inference_conceptual}
\end{figure*}

\subsection{Evidence Representation}

\noindent In order to obtain a measure of absolute evidence by considering a finite number of competing models, we place a Dirichlet distribution over the estimated posterior model probabilities \cite{sensoy2018evidential}. This corresponds to modeling second-order probabilities in terms of the theory of subjective logic (SL) \cite{jsang2018subjective}. These second-order probabilities represent an uncertainty measure over quantities which are themselves probabilities. We use the second-order probabilities to capture epistemic uncertainty about whether the observed data has been generated by one of the candidate models considered during training. 

The probability density function (PDF) of a Dirichlet distribution is given by:
\begin{align}
\textrm{Dir}(\bs{\pi} \given \bs{\alpha}) = \frac{1}{B(\bs{\alpha})} \prod_{j=J}^{J}\pi_{j}^{\alpha_{j} - 1}
\end{align}
where $\bs{\pi}$ belongs to the unit $J-1$ simplex (i.e., $\bs{\pi} \in \Delta^J := \{\bs{\pi} \given \sum_{j=1}^{J}\pi_{j} = 1\}$ and $B(\bs{\alpha})$ is the multivariate beta function. 
The Dirichlet density is parameterized by a vector of \textit{concentration parameters} $\bs{\alpha} \in \mathbb{R}_{+}^{J}$ which can be interpreted as evidences in the ST framework \cite{jsang2018subjective}. 
The sum of the individual evidence components $\alpha_{0} = \sum_{j=1}^{J}\alpha_{j}$ is referred to as the Dirichlet strength, and it affects the precision of the higher-order distribution in terms of its variance. Intuitively, the Dirichlet strength governs the \textit{peakedness} of the distribution, with larger values leading to more peaked densities (i.e., most of the density being concentrated in a smaller region of the simplex).
We can use the mean of the Dirichlet distribution, which is a vector of probabilities given by:
\begin{align}
    \mathbb{E}_{\bs{\pi} \sim \textrm{Dir}(\bs{\alpha})}[\bs{\pi}] = \frac{\bs{\alpha}}{\alpha_0}
\end{align}
to approximate the posterior model probabilities $p(\mathcal{M} \given \x_{1:N})$, as will become clearer later in this section. 
A crucial advantage of such a Dirichlet representation is that it allows to look beyond model probabilities by inspecting the vector of computed evidences. 
For instance, imagine a scenario with three possible models. 
If $\bs{\alpha} = (5,5,5)$, the data provides equally strong evidence for all models (Figure \ref{fig:inference_conceptual}, first column) -- all models explain the data well. 
If, on the other hand, $\bs{\alpha} = (1,1,1)$, then the Dirichlet distribution reduces to a uniform on the simplex indicating no evidence for any of the models (Figure \ref{fig:inference_conceptual}, second column) -- no model explains the observations well. 
Note that in either case one cannot select a model on the basis of the data, because posterior model probabilities are equal, yet the interpretation of the two outcomes is very different:
The second-order Dirichlet distribution allows one to distinguish between \textit{equally likely} (first case) and \textit{equally unlikely} (second case) models. The last column of Figure \ref{fig:inference_conceptual} illustrates a scenario with $\bs{\alpha} = (2, 7, 3)$ in which case one can distinguish between all models (see also \autoref{fig:Fig.9} for a scenario with data simulated from an actual model).

We can further quantify this distinction by computing an uncertainty score given by:
\begin{align}
    u = \frac{J}{\alpha_0} \label{eqn:9} 
\end{align}
\noindent where $J$ is the number of candidate models. 
Importantly, in our framework, individual concentration parameters (resp. neural network outputs) are lower bounded by $1$.
Thus, the uncertainty score ranges between $0$ (total certainty) and $1$ (total uncertainty) and has a straightforward interpretation. Accordingly, total uncertainty is given when $\alpha_0 = J$, which would mean that the data provide no evidence for any of the $J$ candidate models. 
On the other hand, $u << 1$ implies a large Dirichlet strength $\alpha_0 >> J$, which would read that the data provide plenty of evidence for one or more models in question. 
The uncertainty score corresponds to the concept of \textit{vacuity} (i.e., epistemic uncertainty) in the terminology of SL \cite{jsang2018subjective}.
We argue that epistemic uncertainty should be a crucial aspect in model selection, as it quantifies the strength of evidence, and, consequently, the strength of the theoretical conclusions we can draw given the observed data. 

Consequently, model comparison in our framework consists in inferring the parameters of a Dirichlet distribution given an observed dataset. 
The problem of inferring posterior model probabilities can be formulated as:
\begin{align}
    p(\mathcal{M} \given \x_{1:N}) &\approx p_{\bs{\phi}}(\m \given \x_{1:N}) = \mathbb{E}_{\bs{\pi} \sim \textrm{Dir}(f_{\bs{\phi}}(\x_{1:N}))}[\bs{\pi}]  \label{eqn:10}
\end{align}
where $f_{\bs{\phi}}$ is a neural network with positive outputs greater than one, $f_{\phib} : \mathcal{X}^N \rightarrow [1, \infty]^J$. 
Additionally, we can also obtain a measure of absolute model evidence by considering the uncertainty encoded by the full Dirichlet distribution (Eq.\ref{eqn:9}).

\subsection[Learning Evidence in an M-Closed Framework]{Learning Evidence in an $\mathcal{M}$-Closed Framework}

\noindent How do we ensure that the outputs of the neural network match the true unknown model posterior probabilities? 
Consider, for illustrational purposes, a dataset with a single observation, that is $N = 1$ such that $\x_{1:N} = \x$. 
As per \hyperref[alg:1]{\textbf{Algorithm 1}}, we have unlimited access to training samples from $p(\mathcal{M},\x) = \int p(\mathcal{M})p(\thetab \given \mathcal{M})p(\x \given \thetab, \mathcal{M}) d\,\thetab$. 
We use the mean of the Dirichlet distribution $p_{\bs{\phi}}(\m \given \x)$ parameterized by an evidential neural network with parameters $\bs{\phi}$ to approximate $p(\mathcal{M} \given \x)$. 
To optimize the parameters of the neural network, we can minimize some loss $\mathcal{L}$ in expectation over all possible datasets:
\begin{align}
\bs{\phi}^* 
&= \argmin_{\bs{\phi}}\mathbb{E}_{(\m,\x) \sim p(\mathcal{M}, \x)}\left[\mathcal{L}(p_{\bs{\phi}}(\m \given \x),\bs{m})\right] \label{eqn:11}
\end{align}
where $\m$ is a one-hot encoded vector of the true model index $\mathcal{M}_j$. 
We also require that $\mathcal{L}$ be a \textit{strictly proper loss} \cite{gneiting2007strictly}. 
A loss function is strictly proper if and only if it attains its minimum when $p_{\bs{\phi}}(\m \given \x) = p(\mathcal{M} \given \x)$ \cite{gneiting2007strictly}. 
When we choose the Shannon entropy $\mathbb{H}(p_{\bs{\phi}}(\m \given \x))=-\sum_j p_{\bs{\phi}}(\m \given \x)_j \log p_{\bs{\phi}}(\m \given \x)_j$ for $\mathcal{L}$, we obtain the strictly proper logarithmic loss:
\begin{align}
\mathcal{L}(p_{\bs{\phi}}(\m \given \x),\bs{m}) 
&= -\sum_{j=1}^{J} m_{j} \log p_{\bs{\phi}}(\m \given \x)_j \label{eqn:14} \\
&= -\sum_{j=1}^{J} m_{j} \log \left( \frac{f_{\bs{\phi}}(\x)_{j}}{\sum_{j'=1}^{J}f_{\bs{\phi}}(\x)_{j'}} \right) \label{eqn:15}
\end{align}
where $m_j = 1$ when $j$ is the true model index and $0$ otherwise. Thus, in order to estimate $\phi$, we can minimize the expected logarithmic loss over all simulated datasets where $f_{\bs{\phi}}(\x)_{j}$ denotes the $j$-th component of the Dirichlet density given by the evidential neural network. Since we use a strictly proper loss, the evidential network yields the true model posterior probabilities over all possible datasets when perfectly converged.

Intuitively, the logarithmic loss encourages high evidence for the true model and low evidences for the alternative models. Correspondingly, if a dataset with certain characteristics can be generated by different models, evidence for these models will jointly increase. Additionally, the model which generates these characteristics most frequently will accumulate the most evidence and thus be preferred. However, we also want low evidence, or, equivalently, high epistemic uncertainty, for datasets which are implausible under all models. We address this problem in the next section. 

\begin{algorithm*}
\caption{Training phase and inference phase for amortized Bayesian model comparison} \label{alg:2}
\begin{algorithmic}[1]
\Require {$f_{\phib}$ - evidential neural network, $\{\x_{1:N_i}^{(obs)}\}_{i=1}^I$ - list of $I$ observed data sets for inference, $\lambda$ - regularization weight, $B$ - number of simulations at each iteration (batch size).} 
\State \emph{Training phase:}
\Repeat
\State{Generate a training batch $\mathcal{D}^{(B)}_N = \{(\m^{(b)}, \x_{1:N}^{(b)})\}_{b=1}^{B}$ via \hyperref[alg:1]{\textbf{Algorithm 1}}}.
\State{Compute evidences for each simulated data set in $\mathcal{D}^{(B)}_N$: $\bs{\alpha}^{(b)} = f_{\bs{\phi}}(\x_{1:N}^{(b)})$.}
\State {Compute loss according to Eq.\ref{eqn:15}.}
\State {Update neural network parameters $\bs{\phi}$ via backpropagation.}
\Until{convergence to $\widehat{\bs{\phi}}$}
\State \emph{Amortized inference phase:}
\For{$i = 1,...,I$}
\State{Compute model evidences $\bs{\alpha}^{(obs)}_i = f_{\widehat{\bs{\phi}}}(\x_{1:N_{i}}^{(obs)})$.}
\State{Compute uncertainty $u_i = J / \sum_{j=1}^{J}\alpha_{i,j}^{(obs)}$.}
\State{Approximate true model posterior probabilities $p(\mathcal{M} \given \x_{1:N_i}^{(obs)})$ via $p_{\phib}(\m \given \x_{1:N_i}) = \bs{\alpha}_i^{(obs)} / \sum_{j=1}^{J}\alpha_{i,j}^{(obs)}$.} 
\EndFor
\State{Choose further actions.}
\end{algorithmic}
\end{algorithm*}

\subsection{Learning Absolute Evidence through Regularization}

\noindent We now propose a way to address the scenario in which no model explains the observed data well. In this case, we want the evidential network to estimate low evidence for all models in the candidate set. In order to attenuate evidence for datasets which are implausible under all models considered, we incorporate a Kullback-Leibler (KL) divergence into the criterion in Eq.\ref{eqn:14}. We compute the KL divergence between the Dirichlet density generated by the neural network and a uniform Dirichlet density implying total uncertainty. Thus, the KL shrinks evidences which do not contribute to correct model assignments during training, so an implausible dataset at inference time will lead to low evidence under all models. This type of regularization has been used for capturing out-of-distribution (OOD) uncertainty in image classification \cite{sensoy2018evidential}. Accordingly, our modified optimization criterion becomes:
\begin{align}
\bs{\phi}^* = \argmin_{\bs{\phi}}\mathbb{E}_{(\m, \x) \sim p(\mathcal{M}, \x)}\left[ \mathcal{L}(p_{\bs{\phi}}(\m \given \x),\bs{m}) + \lambda \Omega(\Tilde{\bs{\alpha}}) \right] \label{eqn:16} 
\end{align}
with $\Omega(\Tilde{\bs{\alpha}}) = \mathbb{KL}[\textrm{Dir}(\Tilde{\bs{\alpha}}) \,||\, \textrm{Dir}(\bs{1})]$. 
The term $\Tilde{\bs{\alpha}} = \m + (1 - \m) \odot \bs{\alpha}$ represents the estimated evidence vector after removing the evidence for the true model. 
This is possible, because we know the true model during simulation-based training. For application on real data sets after training, knowing the ground truth is not required anymore as $\bs{\phi}$ has been obtained already at this point. The KL penalizes evidences for the false models and drives these evidences towards unity. Equivalently, the KL acts as a ground-truth preserving prior on the higher-order Dirichlet distribution which preserves evidence for the true model and attenuates misleading evidences for the false models. The hyperparameter $\lambda$ controls the weight of regularization and encodes the tolerance of the algorithm to accept implausible (out-of-distribution) datasets during inference. 
With large values of $\lambda$, it becomes possible to
detect cases where all models are deficient; with $\lambda = 0$, only relative evidence is generated. 
Note, that in the latter case, we recover our original proper criterion without penalization. The KL weight $\lambda$ can be selected through prior empirical considerations on how well the simulations cover the plausible set of real-world datasets. 

Importantly, the introduction of the KL regularizer renders the loss no longer \textit{strictly proper}. Therefore, a large regularization weight $\lambda$ would lead to poorer calibration of the approximate model posteriors, as the regularized loss is no longer minimized by the true model posterior. However, since the KL prior is ground-truth preserving, the accuracy of recovering the true model should not be affected. Indeed, we observe this behavior throughout our experiments.

To make optimization tractable, we utilize the fact that we can easily simulate batches of the form $\mathcal{D}^{(B)}_N = \{(\m^{(b)}, \x_{1:N}^{(b)})\}_{b=1}^{B}$ via \hyperref[alg:1]{\textbf{Algorithm 1}} and approximate Eq.\ref{eqn:16} via standard backpropagation by minimizing the following loss:
\begin{align}
    \mathcal{L}(\phib) =  \frac{1}{B}\sum_{b=1}^{B} \left[ -\sum_{j=1}^{J} m_{j}^{(b)} \log \left( \frac{f_{\bs{\phi}}(\x_{1:N}^{(b)})_{j}}{\sum_{j'=1}^{J}f_{\bs{\phi}}(\x_{1:N}^{(b)})_{j'}} \right) + \lambda \Omega(\bs{\tilde{\alpha}}^{(b)}) \right] \label{eqn:15}
\end{align}
over multiple batches to converge at a Monte Carlo estimator $\widehat{\phib}$ of $\phib^*$. In practice, convergence can be determined as the point at which the loss stops decreasing, a criterion similar to \textit{early stopping}. Alternatively, the network can be trained for a pre-defined number of epochs. Note, that, at least in principle, the network can be trained arbitrarily long, since we assume that we can access the joint distribution $p(\mathcal{M}, \bs{x}, N)$ through simulation (cf. Figure \ref{fig:framework}, left panel).

\subsection{Implicit Preference for Simpler Models}

\noindent Remembering that $p_{\phib}(\m \given \x_{1:N}) \propto p(\x_{1:N} \given \mathcal{M})p(\mathcal{M})$, we note that perfect convergence implies that preference for simpler models (Bayesian Occam's razor) is automatically encoded by our method. This is due to the fact that we are approximating an expectation over all possible datasets, parameters, and models. Accordingly, datasets generated by a simpler model tend to be more similar compared to those from a more complex competitor. Therefore, during training, certain datasets which are plausible under both models will be generated more often by the simpler model than by the complex model. Thus, a perfectly converged evidential network will capture this behavior by assigning higher posterior probability to the simpler model (assuming equal prior probabilities). Therefore, at least in theory, our method captures complexity differences arising purely from the generative behavior of the models and does not presuppose an \textit{ad hoc} measure of complexity (e.g., number of parameters).

\subsection{Putting it All Together}

\noindent The essential steps of our evidential method are summarized in \hyperref[alg:2]{\textbf{Algorithm 2}}. Note, that steps 2-7 and 9-13 can be executed in parallel with GPU support in order to dramatically accelerate convergence and inference. 
In sum, we propose to cast the problem of model selection as evidence estimation and learn a Dirichlet distribution over posterior model probabilities directly via simulations from the competing models. 
To this end, we train an evidential neural network which approximates posterior model probabilities and further quantifies the epistemic uncertainty as to whether an observed data set is within the generative scope of the candidate models.
Moreover, once trained on simulations from a set of models, the network can be reused and extended with new models across a research domain, essentially \textit{amortizing} the model comparison process. Accordingly, if the priors over model parameter do not change, multiple researchers can reuse the same network for multiple applications. 
If the priors over model parameters change or additional models need to be considered, the parameters of a pre-trained network can be adjusted or the network extended with additional output nodes for the new models.

\section{Experiments}

\noindent In this section, we demonstrate the utility of our method on a toy example and relevant models from chemistry, cognitive science and neurobiology. 
A further toy example with $400$ models as well as details for neural network training, architectures, performance metrics, and forward models are to be found in the \textbf{Appendix}.

\subsection{Experiment 1: Beta-Binomial Model with Known Analytical Marginal Likelihood}

\begin{figure}
\centering
\begin{subfigure}{.24\textwidth}
    \includegraphics[width=\textwidth]{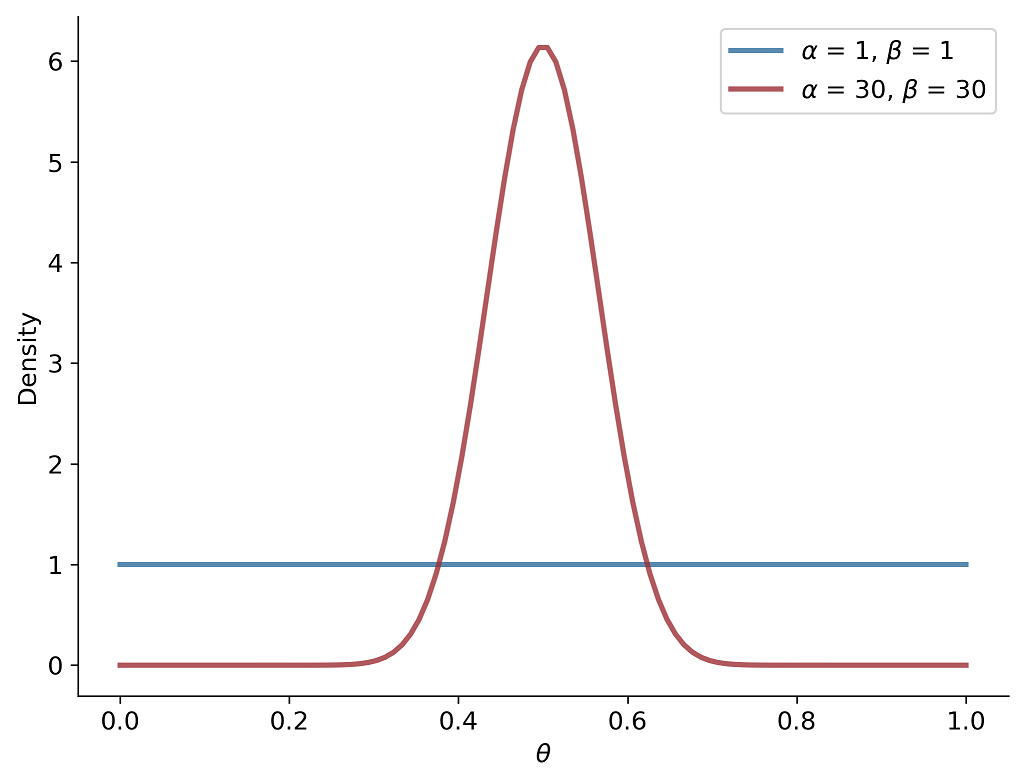}
    \caption{Models' priors}
    \label{fig:Fig.4a}
\end{subfigure}
\begin{subfigure}{.24\textwidth}
    \includegraphics[width=\textwidth]{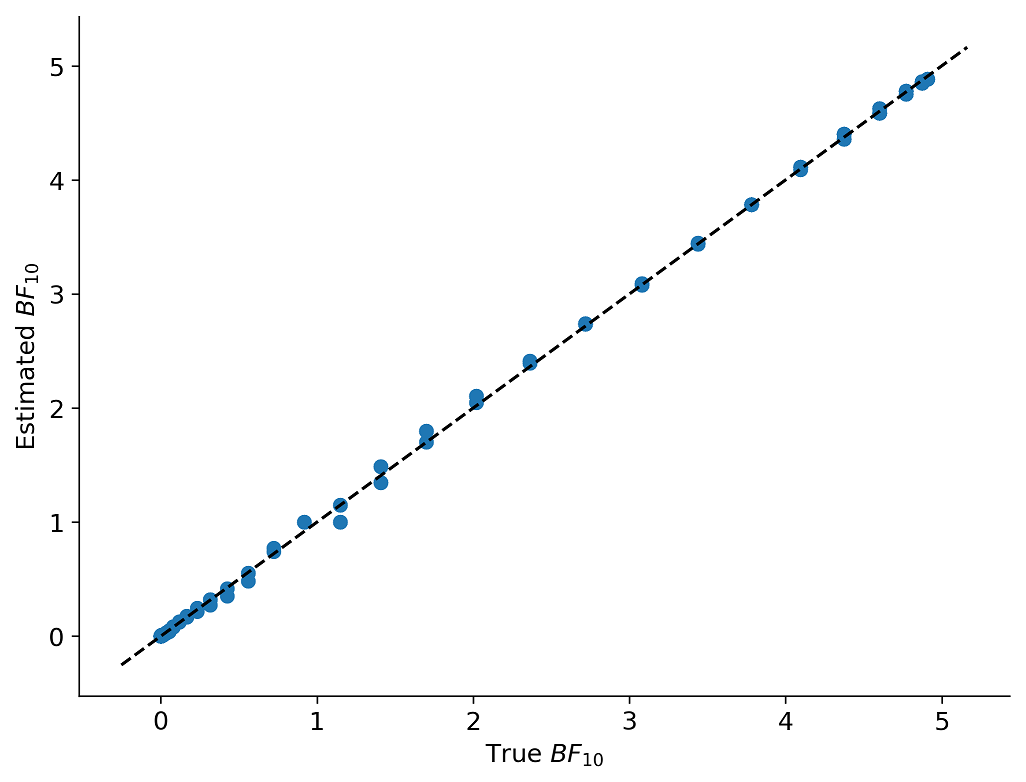}
    \caption{True vs. estimated Bayes factors at $N=100$}
    \label{fig:Fig.4b}
\end{subfigure}
\begin{subfigure}{.24\textwidth}
    \includegraphics[width=\textwidth]{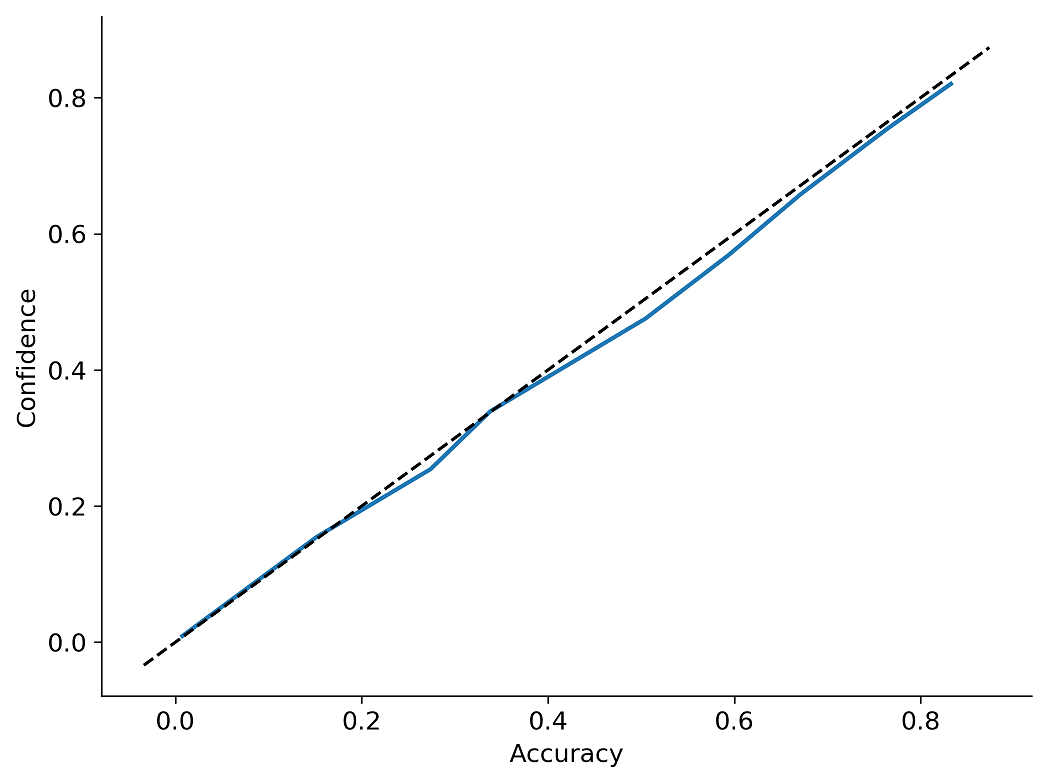}
    \caption{Calibration curve at $N=100$}
    \label{fig:Fig.4c}
\end{subfigure}
\begin{subfigure}{.24\textwidth}
    \includegraphics[width=\textwidth]{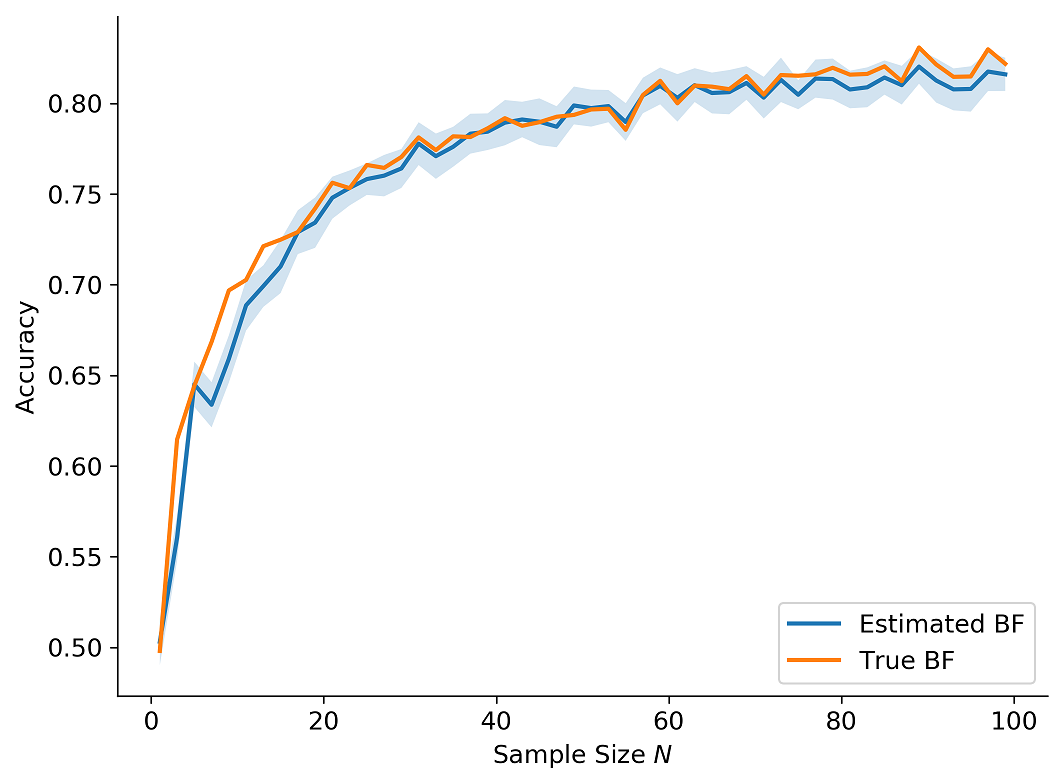}
    \caption{Accuracy across all $N$}
    \label{fig:Fig.4d}
\end{subfigure}
\caption[short]{\textbf{(a)} Prior densities of the theta parameter for both models of \textbf{Experiment 1}; \textbf{(b)} True vs. estimated Bayes factors obtained from the network-induced Dirichlet distribution at $N=100$; \textbf{(c)} Calibration curve at $N=100$ indicating very good calibration (dotted line represents perfect calibration); \textbf{(d)} Accuracy at all $N$ achieved with both the analytic and the estimated Bayes factors (the shaded region represents a 95\% bootstrap confidence interval around the accuracies of the evidential network).} \label{fig:Fig.4}
\end{figure}

\noindent As a basic proof-of-concept for our evidential method, we apply it to a toy model comparison scenario with an analytically tractable marginal likelihood. Here, we pursue the following goals. 
First, we want to demonstrate that the estimated posterior probabilities closely approximate the analytic model posteriors. 
To show this, we compare the analytically computed vs. the estimated Bayes factors. 
In addition, we want to show that accuracy of true model recovery matches closely the accuracy obtained by the analytic Bayes factors across all $N$. For this, we consider the simple beta-binomial model given by:
\begin{align}
    \theta &\sim \textrm{Beta}(\alpha, \beta) \\
    x_j &\sim \textrm{Bernoulli}(\theta) \textrm{ for } j = 1,...,N  
\end{align}
The analytical marginal likelihood of the beta-binomial model is:
\begin{align}
    p(x_{1:N}) = \binom{N}{K}\frac{\textrm{Beta}(\alpha + K, \beta + N - K)}{\textrm{Beta}(\alpha, \beta)}
\end{align}
where $K$ denotes the number of successes in the $N$ trials. 
For this example, we will consider a model comparison scenario with two models, one with a flat prior $\textrm{Beta}(1,1)$ on the parameter $\theta$, and one with a sharp prior $\textrm{Beta}(30,30)$. The two prior densities are depicted in \autoref{fig:Fig.4a}.

We train a small permutation invariant evidential network with batches of size $B=64$ until convergence. For each batch, we draw the samples size from a discrete uniform distribution $N \sim \mathcal{U}_D(1, 100)$ and input the raw binary data to the network. We validate the network on 5000 separate validation datasets for each $N$. Convergence took approximately 15 minutes, whereas inference on all 5000 validation datasets took less than 2 seconds. 

Our results demonstrate that the estimated Bayes factors closely approximate the analytic Bayes factors (\autoref{fig:Fig.4b}). 
We also observe no systematic under- or overconfidence in the estimated Bayes factors, which is indicated by the calibration curve resembling a straight line (\autoref{fig:Fig.4c}). Finally, the accuracy of recovery achieved with the estimated Bayes factors closely matches that of the analytic Bayes factors across all sample sizes $N$ (\autoref{fig:Fig.4d}).

\subsection{Experiment 2: Markov Jump Process of Stochastic Chemical Reaction Kinetics}

\noindent In this experiment, we apply our evidential method to a simple model of non-exchangeable chemical molecule concentration data. 
Further, we demonstrate the efficiency benefits of our amortized learning method compared to the standard non-amortized ABC-SMC algorithm. 

We define two Markov jump process models $\mathcal{M}_1$ and $\mathcal{M}_2$ for conversion of (chemical) species $z$ to species $y$:
\begin{align}
    \mathcal{M}_1&: z + y \xrightarrow{\theta_1} 2y \\ 
    \mathcal{M}_2&: z \xrightarrow{\theta_2} y
\end{align}

Each model has a single rate parameter $\theta_i$. We use the Gillespie simulator to generate simulated time-series from the two models with an upper time limit of $0.1$ seconds. Both models start with initial concentrations $x_0 = 40$ and $y_0 = 3$, so they only differ in terms of their reaction kinetics. The input time-series $\bs{x}_{1:N}$ consist of a time vector $t_{1:N}$ as well as two vectors of molecule concentrations for each species at each time step, $z_{1:N}$ and $y_{1:N}$, which we stack together. We place a wide uniform prior over each rate parameter: $\theta_i \sim \mathcal{U}(0, 100)$.

\begin{figure}
\centering
\begin{subfigure}{.49\textwidth}
    \includegraphics[width=\textwidth]{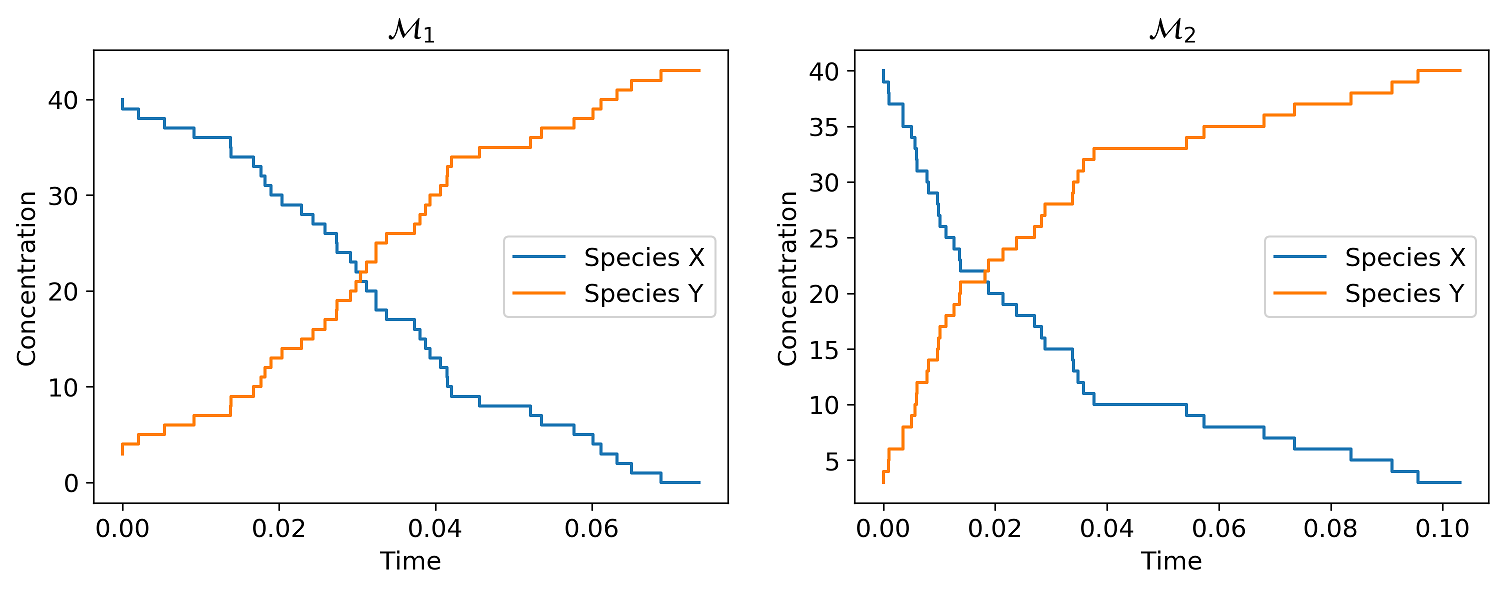}
\end{subfigure}
\caption{Observed concentration time-series from both Markov jump models of \textbf{Experiment 2} with $\theta = 2.0$.} \label{fig:Fig.6}
\end{figure}

We train an evidential sequence network for $50$ epochs of $1000$ mini-batch updates and validate its performance on $500$ previously unobserved time-series. Wall-clock training time was approximately $52.3$ minutes. In contrast, wall-clock inference time on the $500$ validation time-series was $254$ ms, leading to dramatic amortization gains. The bootstrap accuracy of recovery was $0.98\ (SD=0.01)$ over the entire validation set.

We also apply the ABC-SMC algorithm available from the \textit{pyABC} \cite{klinger2018pyabc} library to a single data-set $\x_{1:N}^{(obs)}$ generated from model 1 ($\mathcal{M}_1$) with rate parameter $\theta_1 = 2.0$. \autoref{fig:Fig.6} depicts the observed data generated from model 1 (left panel) as well as observed data generated from model 2 with $\theta_2 = 2.0$. Notably, the differences in the data-generating processes defined by the two models are subtle and not straightforward to explicitly quantify.

For the ABC-SMC method, we set the minimum rejection threshold $\epsilon$ to $0.7$ and the maximum number of populations to $15$, as these settings lead to perfect recovery of the true model. As a distance function, we use the $L_2$ norm between the raw concentration times-series of species $z$, evaluated at $20$ time points.\footnote{These settings were picked from the original pyABC documentation available at \url{https://pyabc.readthedocs.io/en/latest/examples/chemical_reaction.html}}

The convergence of the ABC-SMC algorithm on the single data set took $12.2$ minutes wall-clock time. Thus, inference on the $500$ validation data-sets would have taken more than $4$ days to complete. Accordingly, we see that the training effort with our method is worthwhile even after as few as 5 data-sets. As for recovery on the single test dataset, ABC-SMC selects the true model with a probability of $1$, whereas our evidential networks outputs a probability of $0.997$ which results in a negligible difference of $0.003$ between the results from two methods.

\subsection{Experiment 3: Stochastic Models of Decision Making}

\noindent In this experiment, we apply our evidential method to compare several non-trivial nested stochastic \textit{evidence accumulator models} (EAMs) from the field of human decision making \cite{usher2001time, ratcliff2008diffusion}. 
With this experiment, we want to demonstrate the performance of our method in terms of accuracy and posterior calibration on exchangeable data obtained from complex cognitive models. 
Additionally, we want to demonstrate how our regularization scheme can be used to capture absolute evidence by artificially rendering the data implausible under all models.

\subsubsection{Model Comparison Setting}

EAMs describe the dynamics of decision making via different neurocognitively plausible parameters (i.e., speed of information processing, decision threshold, bias/pre-activation, etc.). 
EAMs are most often applied to choice reaction times (RT) data to infer neurocognitive processes underlying generation of RT distributions in cognitive tasks. The most general form of an EAM is given by a stochastic differential equation:

\begin{align} 
dx = vdt + cd\xi  
\end{align}

\noindent where $dx$ denotes a change in activation of an accumulator, $v$ denotes the average speed of information accumulation (often termed the drift rate), and $d\xi$ represents a stochastic additive component with $d\xi \sim \mathcal{N}(0, c^{2})$.

Multiple flavors of the above stated basic EAM form exist throughout the literature \cite{usher2001time, ratcliff2008diffusion, turner2016bayesian, brown2008simplest}. Moreover, most EAMs are intractable with standard Bayesian methods \cite{brown2008simplest}, so model selection is usually hard and computationally cumbersome. With this example, we pursue several goals. First, we want to demonstrate the utility of our method for performing model selection on multiple nested models. Second, we want to empirically show that our method implements Occam's razor. Third, we want to show that our method can indeed provide a proxy for absolute evidence.

To this end, we start with a very basic EAM defined by four parameters $\thetab = (v_1, v_2, a, t_0)$ with $v_i$ denoting the speed of information processing (drift rate) for two simulated RT experimental tasks $i \in \{1, 2\}$, $a$ denoting the decision threshold, and $t_0$ denoting an additive constant representing the time required for non-decisional processes like motor reactions. We then define five more models with increasing complexity by successively \textit{freeing} the parameters $z_r$ (bias), $\alpha$ (heavy-taildness of noise distribution), $s_{t_{0}}$ (variability of non-decision time), $s_v$ (varibaility of drift-rate), and $s_{zr}$ (variability of bias). Note, that the inclusion of non-Gaussian diffusion noise renders an EAM model intractable, since in this case no closed-form likelihood is available (see \cite{voss2019sequential} for more details). \autoref{table:Table 1} lists the priors over model parameters as well as fixed parameter values.

The task of model selection is thus to choose among six nested EAM models $\mathcal{M} = \{\mathcal{M}_1, \mathcal{M}_2, \mathcal{M}_3, \mathcal{M}_4, \mathcal{M}_5, \mathcal{M}_6\}$, each able to capture increasingly complex behavioral patterns. 
Each model $j$ is able to account for all datasets generated by the previous models $i < j$, since the previous models are nested within the $j$-th model. For instance, model $\mathcal{M}_6$ can generate all data sets possible under the other models at the cost of increased functional and parametric complexity. Therefore, we need to show that our method encodes Occam's razor purely through the generative behavior of the models. 

In order to show that our regularization method can be used as a proxy to capture absolute evidence, we perform the following experiment. We define a temporal shifting constant $K \in (0, 10)$ (in units of seconds) and apply the shift to each response time in each validation data set. Therefore, as $K$ increases, each data set becomes increasingly implausible under all models considered. For each $K$, we compute the average uncertainty over all shifted validation data sets and plot is as a function of $K$. Here, we only consider the maximum number of trials $N = 300$.

We train three evidential neural networks with different KL weights: $\lambda \in \{0.0, 0.1, 1.0\}$ in order to investigate the effects of $\lambda$ on accuracy, calibration, and uncertainty. All networks were trained with variable number of trials $N \sim \mathcal{U}_D(1, 300)$ per batch for a total of 50000 iterations. The training of each network took approximately half a day on a single computer. In contrast, performing inference on 5000 data sets with a pre-trained network took less than 10 seconds.

\begin{figure*}
\centering
\begin{subfigure}{.32\textwidth}
    \includegraphics[width=\textwidth]{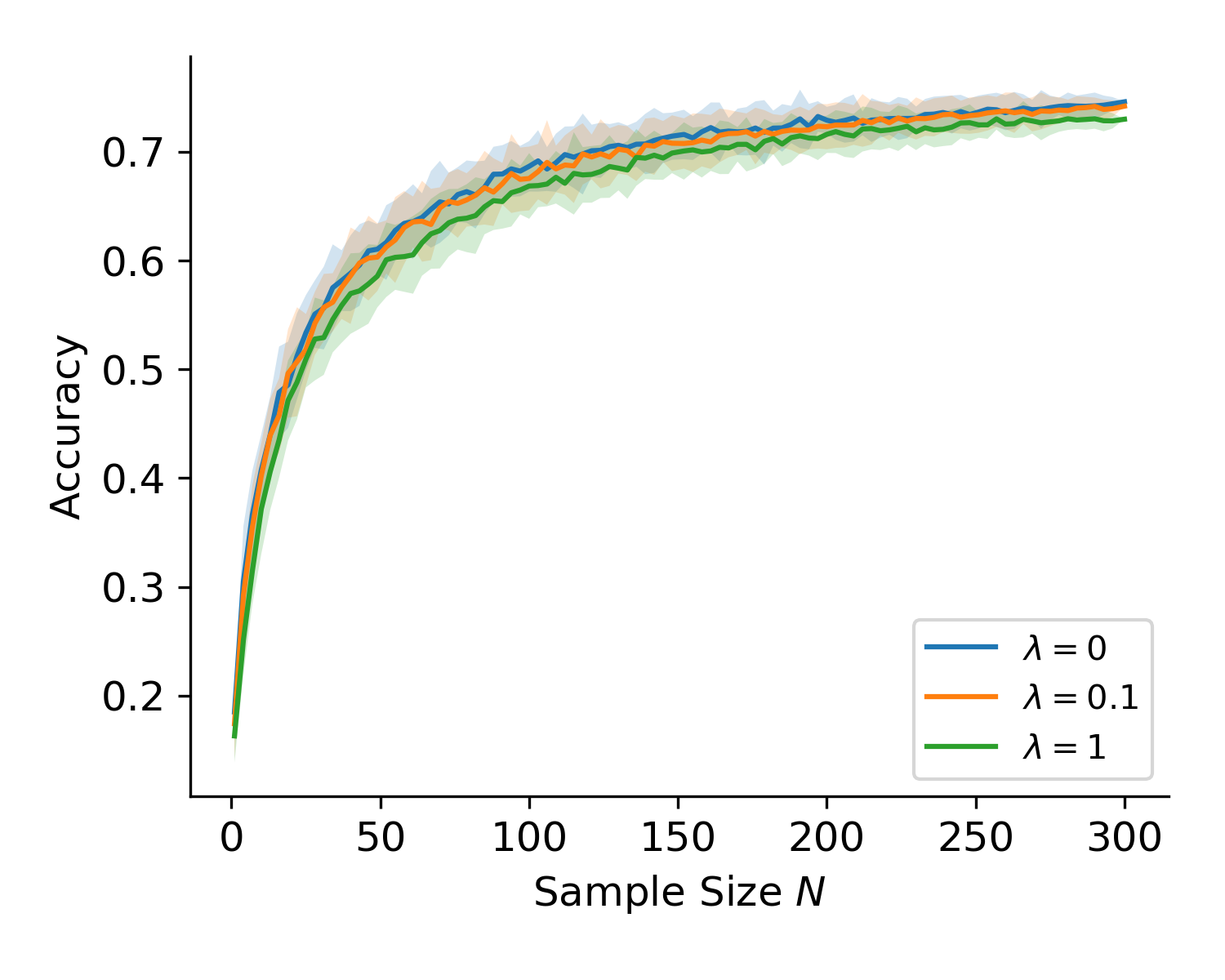}
    \caption{Accuracy over all $N$}
    \label{fig:Fig.7a}
\end{subfigure}
\begin{subfigure}{.32\textwidth}
    \includegraphics[width=\textwidth]{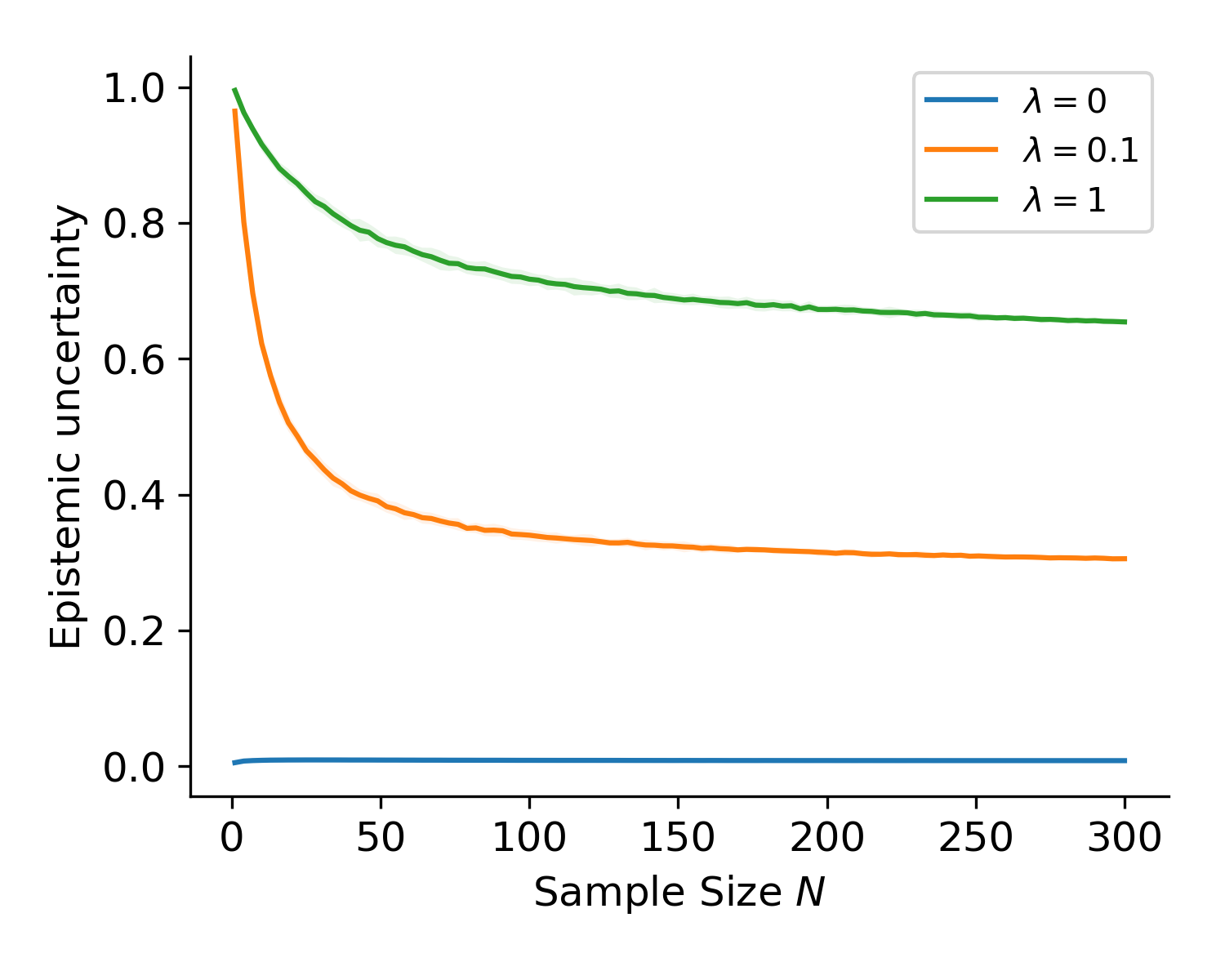}
    \caption{Epistemic uncertainty over all $N$}
    \label{fig:Fig.7b}
\end{subfigure}
\begin{subfigure}{.32\textwidth}
    \includegraphics[width=\textwidth]{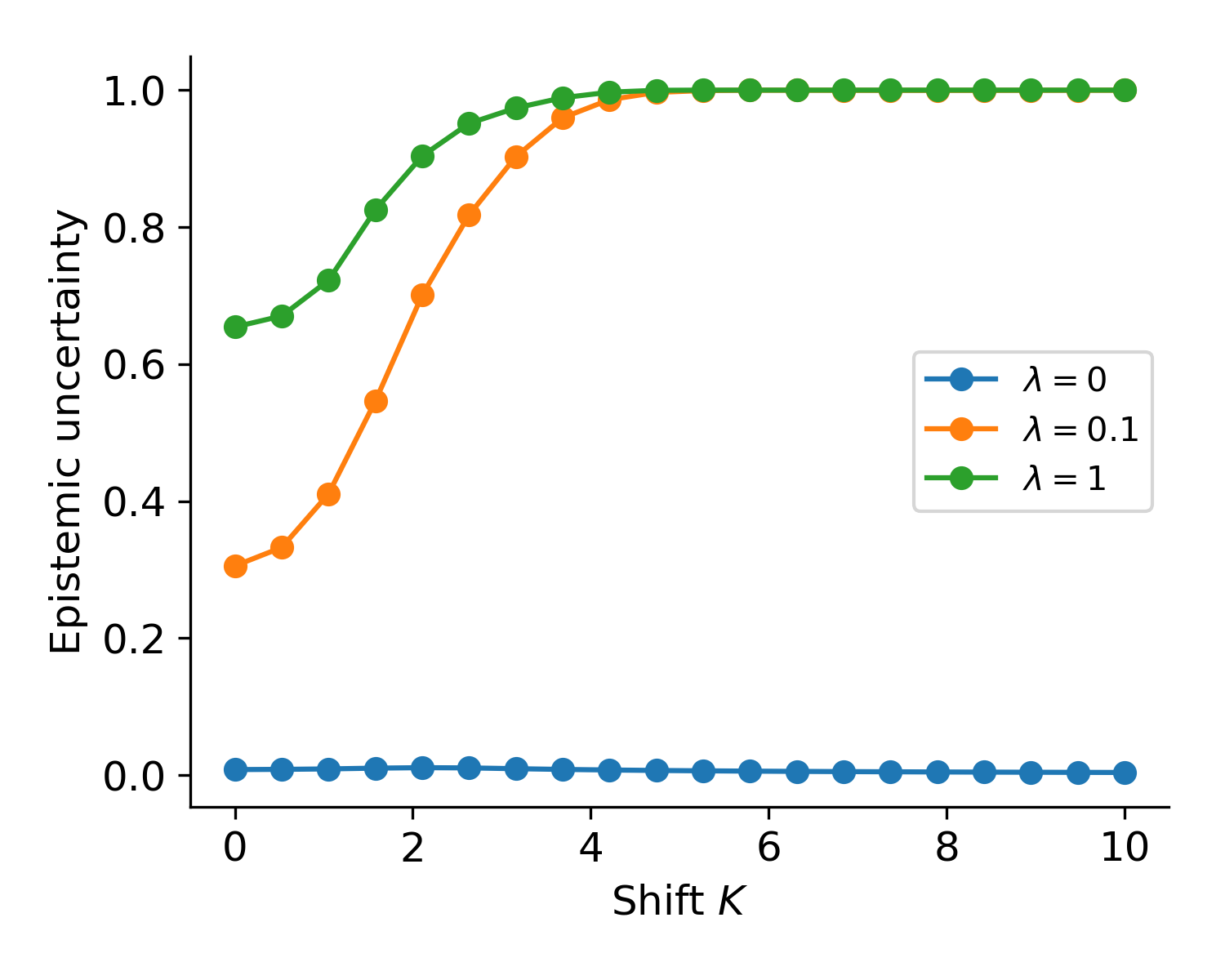}
    \caption{Epistemic uncertainty over all $K$}
    \label{fig:Fig.7c}
\end{subfigure}
\begin{subfigure}{.46\textwidth}
    \includegraphics[width=\textwidth]{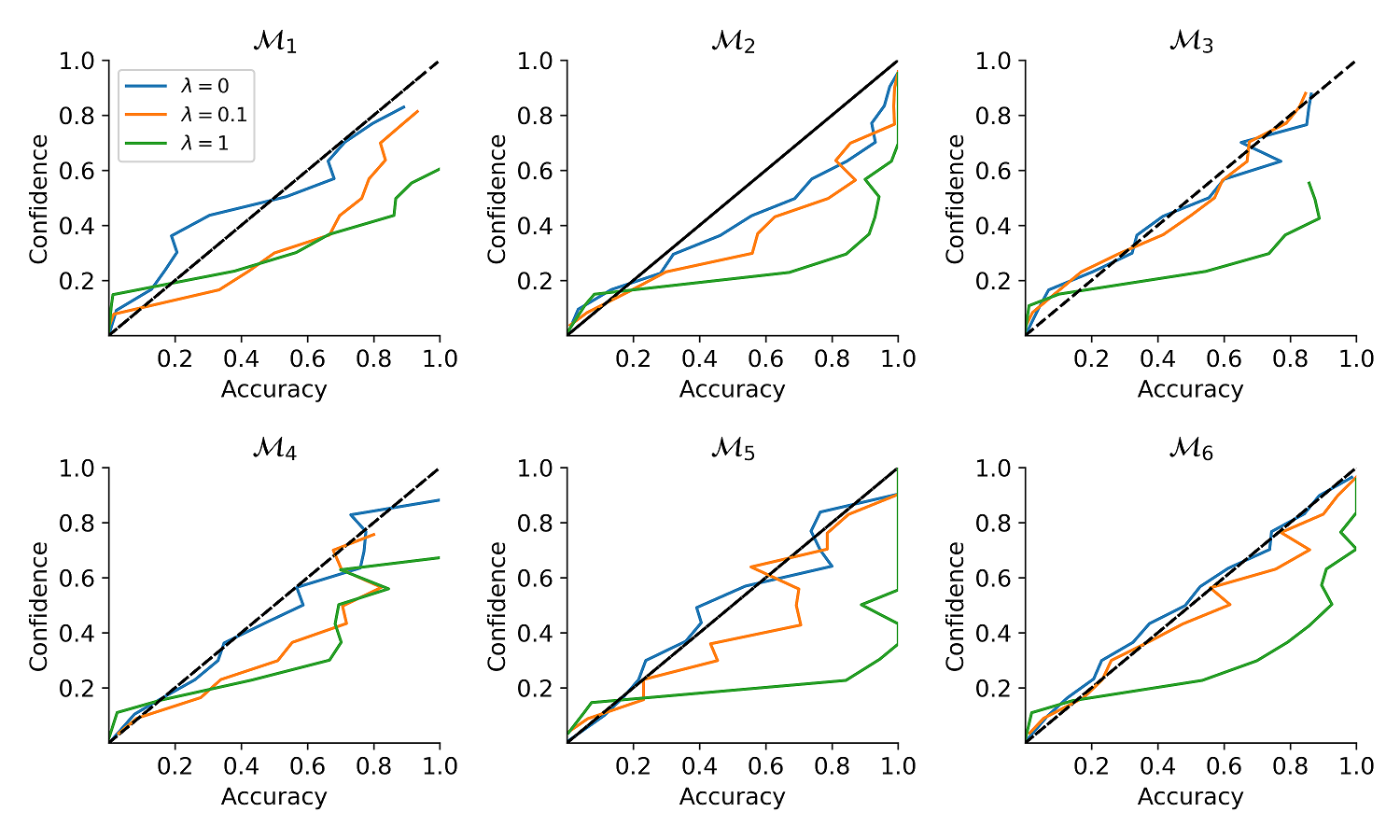}
    \caption{Calibration curves at $N=300$}
    \label{fig:Fig.7d}
\end{subfigure}
\begin{subfigure}{.46\textwidth}
    \includegraphics[width=\textwidth]{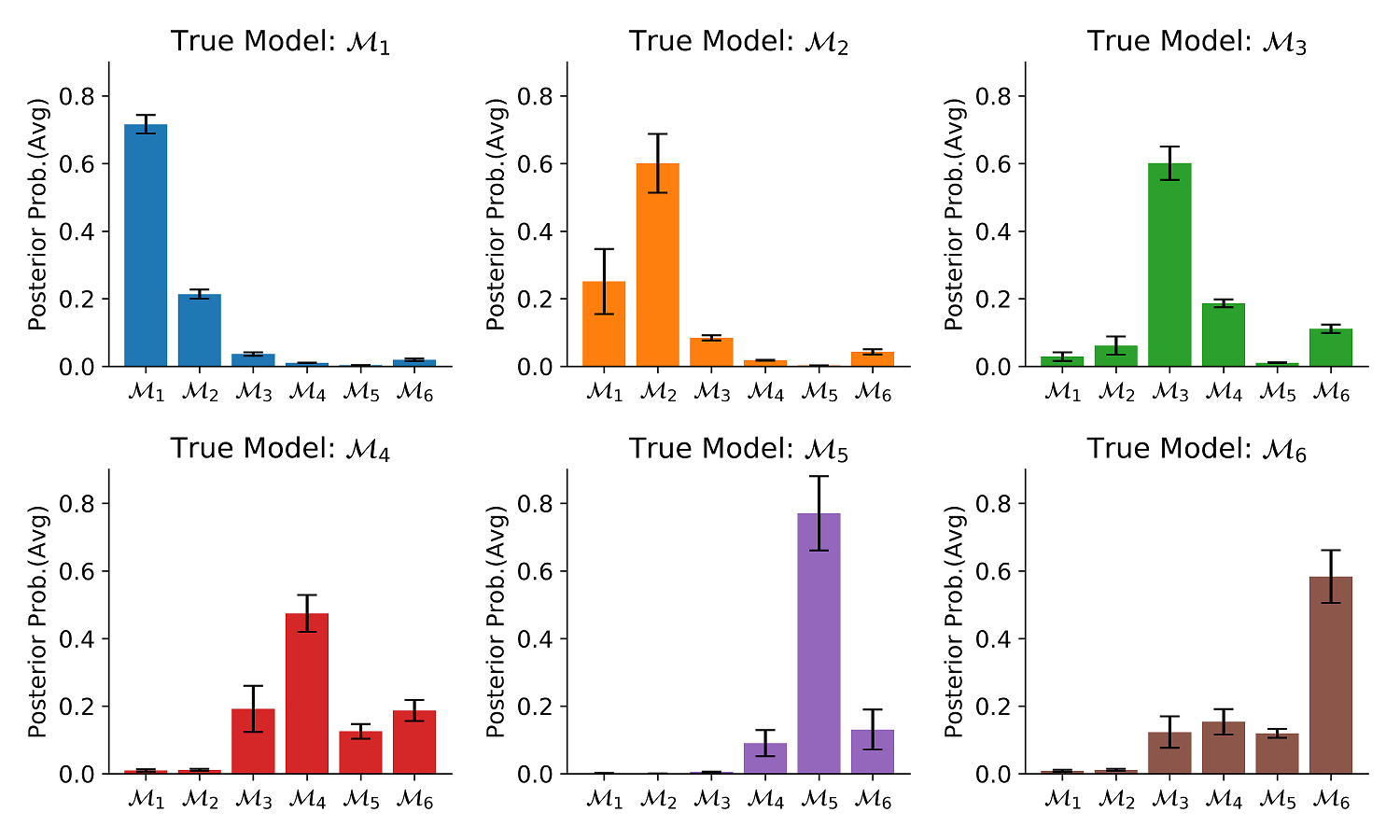}
    \caption{Occam's razor at $N=300$}
    \label{fig:Fig.7e}
\end{subfigure}
\caption{Detailed validation results from \textbf{Experiment 3}.}\label{fig:Fig.7}
\end{figure*}

\subsubsection{Validation Results}

To quantify the global performance of our method, we compute the accuracy of recovery as a function of the number of observations ($N$) for each of the models. We also compute the epistemic uncertainty as a function of $N$. To this end, we generate 500 new datasets for each $N$ and compute the accuracy of recovery and average uncertainty. These results are depicted in \autoref{fig:Fig.7}. 

\textit{Accuracy}. We observe that accuracy of recovery increases with increasing sample size and begins to flatten out around $N=100$, independently of the regularization weight $\lambda$ (\autoref{fig:Fig.7a}). This behavior is desirable, as selecting the true model should become easier when more information is available. Further, since the models are nested, perfect recovery is not possible, as the models exhibit a large shared data space.

\textit{Calibration}. \autoref{fig:Fig.7d} depicts calibration curves for each model and each regularization value. The unregularized network appears to be very well calibrated, whereas the regularized networks become increasingly underconfident with increasing regularization weight. This is due to the fact that the regularized networks are encouraged to generate zero evidence for the wrong models, so model probabilities become miscalibrated. Importantly, none of the networks shows overconfidence.

\textit{Occam's Razor}. We also test Occam's razor by generating $500$ data sets from each model with $N=300$ and compute the average predicted model posterior probabilities by the unregularized network. Thus, all data sets generated by model $j$ are plausible under the remaining models $\mathcal{M}_i, i > j$. These average model probabilities are depicted in \autoref{fig:Fig.7e}. Even though data-set generated by the nested simpler models are plausible under the more complex models, we observe that Occam's razor is encoded by the behavior of the network, which, on average, consistently selects the simpler model when it is the true data-generating model. We also observe that this behavior is independent of regularization (results for $\lambda=0.1$ and $\lambda = 1$ are not depicted in \autoref{fig:Fig.7e})

\textit{Epistemic uncertainty and absolute evidence}. Epistemic uncertainty over different trial numbers ($N$) is zero when no KL regularization is applied ($\lambda = 0$). On the other hand, both small ($\lambda = 0.1$) or large ($\lambda=1.0$) regularization weights lead to non-zero uncertainty over all possible $N$ (\autoref{fig:Fig.7b}). This pattern reflects a reduction in epistemic uncertainty with increasing amount of information and mirrors the inverse of the recovery curve. Note, that the value at which epistemic uncertainty begins to flatten out is larger for the highly regularized model, as it encodes more cautiousness with respect to the challenging task of selecting a true nested model. Finally, results on shifted data-sets are depicted in \autoref{fig:Fig.7c}. Indeed, we observe that the regularized networks are able to detect implausible data sets and output total uncertainty around $K>4$ for all manipulated data sets. Uncertainty increases faster for high regularization. On the other hand, the unregularized model does not have any way of signaling impossibility of a decision, so its uncertainty remains at $0$ over all $K$.

\subsection{Experiment 4: Stochastic Models of Single-Neuron Activity}

\begin{figure*}
\centering
\begin{subfigure}{.9\textwidth}
    \includegraphics[width=\textwidth]{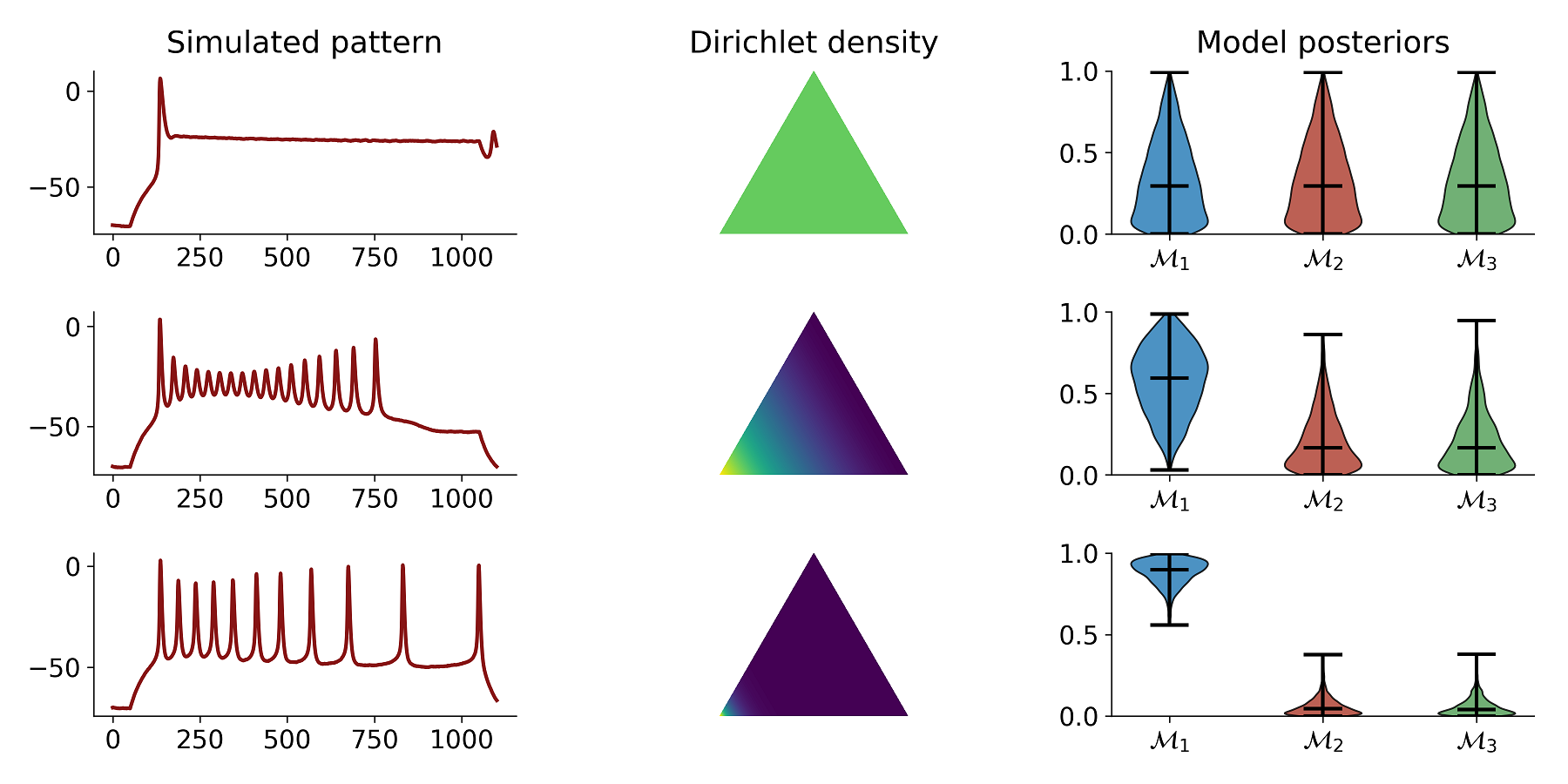}
\end{subfigure}
\caption{Three simulated firing patterns, corresponding estimated Dirichlet densities and model posteriors. Each row illustrates a different value of the parameter $\bar{g}_K$: $\bar{g}_K = 0.1$, $\bar{g}_K = 0.5$, and $\bar{g}_K = 0.75$, respectively. An increase in the parameter $\bar{g}_K$ is accompanied by a decrease in epistemic uncertainty (as measured via Eq.\ref{eqn:9}). An implausible value of $\bar{g}_k$ (first row) results in a flat density as an index of total uncertainty (uniform green areas). As the parameter value surpasses the plausible boundary (second and third rows), the Dirichlet simplex becomes peaked towards the lower left edge encoding $\mathcal{M}_1$.}
\label{fig:Fig.9}
\end{figure*}

\noindent In this experiment, we apply our evidential method to complex nested spiking neuron models describing the properties of biological cells in the nervous system. 
The purpose of this experiment is threefold. 
First, we want to assess the ability of our method to classify models deploying a variety of spiking patterns which might account for different cortical and sub-cortical neuronal activity. 
Second, we want to challenge the network's ability to detect biologically implausible data patterns as accounted by epistemic uncertainty. 
Finally, we compare our method with other viable neural network architectures that are able to perform amortized model comparison as classification. 
To this aim, we rely on a renowned computational model of biological neuronal dynamics.

\subsubsection{Model Comparison Setting}

In computational neuroscience, mathematical modeling of neuroelectric dynamics serves as a basis to understand the functional organization of the brain from both single-cell and large-scale network processing perspectives \cite{hodgkin1952quantitative,burkitt2006review,izhikevich2003simple,abbott1990model,dura2019netpyne}. 
A plurality of different neural models have been proposed during the last decades, spanning from completely abstract to biologically detailed models. 
The former offer a simplified mathematical representation able to account for the main functional properties of spiking neurons, the latter provide a detailed analogy between models' state variables and ion channels in biological neurons \cite{pospischil2011comparison}. 
Importantly, these computational models differ in their capability to reproduce firing patterns observed in real cortical neurons \cite{izhikevich2004model}.

The model family we consider here is a Hodgkin-Huxley type model of cerebral cortex and thalamic neurons \cite{pospischil2008minimal,hodgkin1952quantitative}. 
The forward model is formulated as a set of five ordinary differential equations (ODEs) describing how the neuron membrane potential $V(t)$ unfolds in time as a function of an injected current $I_{inj}(t)$, and ion channels properties.
See \textbf{Appendix D} for more details regarding the forward simulation process.

To set up the model comparison problem, we treat different types of conductance, $g_L,\Bar{g}_{Na},\Bar{g}_K$ and $\Bar{g}_M$, as free parameters, and formulate three neural models based on different parameter configurations. 
In particular, we consider three models $\mathcal{M} = \{\mathcal{M}_1, \mathcal{M}_2, \mathcal{M}_3\}$ defined by the parameter sets $\thetab_1 = (\Bar{g}_{Na},\Bar{g}_K)$, $\thetab_2 = (\Bar{g}_{Na},\Bar{g}_K,\Bar{g}_M)$, and $\thetab_3 = (\Bar{g}_{Na},\Bar{g}_K,\Bar{g}_M,g_L)$. 
When not treated as free parameters, we set $\Bar{g}_M$ and $g_L$ to default values, such that $\Bar{g}_M=0.07$ and $g_L=0.1$.

We compare the performance of our evidential method to the following methods: a standard softmax classifier, a classifier with Monte Carlo dropout \cite{gal2016dropout}, and two variational classifiers using a Kullback-Leibler \cite{kingma2015variational} and a maximum mean discrepancy (MMD, \cite{zhao2017infovae}) latent space regularizer, respectively. 
We also train three evidential networks with $\lambda=0$ (no regularization), $\lambda=0.5$, and $\lambda=1.0$ to better quantify the effects of performing regularized model comparison.
Here, we do not consider non-amortized methods, such as ABC or ABC-MCMC, as implemented in \cite{mertens2018abrox}, since they would have taken an infeasible amount of time to validate on hundreds of data sets.  

\subsubsection{Validation Results}

In order to assess performance, we train an unregularized evidential network for $60$ epochs resulting in $60000$ mini-batch updates. 
For each batch, we draw a random input current duration $T \sim \mathcal{U}_D(100, 400)$ (in units of milliseconds), with the same constant input current, $I_{inj}$, for each dataset simulation. 
Here, $T$ reflects the physical time window in which biological spiking patterns can unfold. 
Since the sampling rate of membrane potential is fixed ($dt=0.2$), $T$ affects both the span of observable spiking behaviour and the number of simulated data points.
The entire training phase with online learning took approximately $2.5$ hours. 
On the other hand, model comparison on $5000$ validation time series took approximately $0.7$ seconds, which highlights the extreme efficiency gains obtainable via globally amortized inference.

Regarding model selection, we observe accuracies above $0.92$ across all $T$, with no gains in accuracy for increasing $T$, which shows that even short input currents are sufficient for performing reliable model selection for these complex models. 
Further, mean bootstrap calibration curves and accuracies on $5000$ validation data sets are depicted in \autoref{fig:Fig.8d}.
We observe good calibration for all three models, with calibration errors less than $0.1$. Notably, overconfidence was $0$ for all three models. The normalized \textit{confusion matrix} is depicted in \autoref{fig:Fig.8b}. 

\begin{table}
\centering
\caption{Comparison results from \textbf{Experiment 4}}
\begin{tabular}{lll}
\toprule
Neural Architecture &           Accuracy &        Calibration Error  \\
\midrule
Evidential ($\lambda=0$)   &  $0.919 \pm 0.004$ &  $\bs{0.078 \pm 0.010}$\\
Evidential ($\lambda=0.5$) &  $0.917 \pm 0.006$ &  $0.097 \pm 0.012$\\
Evidential ($\lambda=1.0$) &  $0.900 \pm 0.006$ &  $0.095 \pm 0.011$\\
Softmax classifier       &  $0.913 \pm 0.006$ &  $0.105 \pm 0.015$\\
MC Dropout classifier       &  $0.885 \pm 0.005$ &  $0.087 \pm 0.009$\\
MMD-VAE classifier       &  $0.906 \pm 0.006$ &  $0.091 \pm 0.012$\\
VAE classifier           &  $\bs{0.924 \pm 0.005}$ &  $0.096 \pm 0.012$\\
\bottomrule
\label{tab:comp}
\end{tabular}
\end{table}

In order to asses how well we can capture epistemic uncertainty for biologically implausible firing patterns, we train another evidential network with a gradually increasing regularization weight up to $\lambda = 1.0$. 
We then fix the parameter $\bar{g}_{Na} = 4.0$ of model $\mathcal{M}_1$ and gradually increase its second parameter $\bar{g}_K$ from $0.1$ to $2.0$. Since spiking patterns observed with low values of $\bar{g}_K$ are quite implausible and have not been observed during training, we expect uncertainty to gradually decrease. Indeed, \autoref{fig:Fig.8c} shows this pattern. Three example firing patterns and the corresponding posterior estimates are depicted in \autoref{fig:Fig.9}. On the other hand, changing the sign of the output membrane potential, which results in biologically implausible firing patterns, leads to a trivial selection of $\mathcal{M}_3$. This is contrary to expectations, and shows that absolute evidence is also relative to what the evidential network has learned during training.

Finally, Table \ref{tab:comp} presents the comparison results in terms of accuracy and calibration error (all methods achieved $0$ overconfidence).
We train each neural network method for $30$ epochs with identical optimizer settings and the same recurrent network architecture for ease of comparison.
We then compute validation metrics on $3000$ simulated neural firing patterns and report means and standard errors.
Our unregularized evidential network ($\lambda=0$) achieves the lowest calibration error, followed by the MC dropout classifier.
In terms of accuracy, the KL variational classifier performs slightly better than our unregularized evidential network (but still within one standard error).
Overall, the performance of all amortized methods considered in this experiment is similar, which highlights the viability of the approach to Bayesian model comparison in general. 
Note, that training of each method took less than $1.5$ hours, and bootstrap validation on $3000$ less than a minute. 
The latter would have been impossible to achieve within a reasonable time-frame using non-amortized methods.

\section{Discussion}

\noindent In the current work, we introduced a novel simulation-based method for approximate Bayesian model comparison based on specialized evidential neural networks.
We demonstrated that our method can successfully deal with both exchangeable and non-exchangeable (time-dependent) sequences with variable numbers of observations without relying on fixed summary statistics.
Further, we presented a way to amortize the process of model comparison for a given family of models by splitting it into a potentially costly global training phase and a cheap inference phase.
In this way, pre-trained evidential networks can be stored, shared, and reused across multiple data sets and model comparison applications.
Finally, we demonstrated a way to obtain a measure of absolute evidence in spite of operating in an $\mathcal{M}$-closed framework during the simulation phase. In the following, we reiterate the main advantages of our method.

\textit{Theoretical guarantee}. 
By using a strictly proper loss \cite{gneiting2007strictly}, we showed that our method can closely approximate analytic model posterior probabilities and Bayes factors in theory and practice. In other words, posterior probability estimates are perfectly calibrated to the true model posterior probabilities when the strictly proper logarithmic loss is globally minimized. Indeed, our experiments confirm that the network outputs are well calibrated. However, when optimizing the regularized version of the logarithmic loss, we are no longer working with a strictly proper loss, so calibration declines at the cost of capturing implausible data sets. However, we demonstrated that the accuracy of recovery (i.e., selecting the most plausible model in the set of considered models) does not suffer when training with regularization. In any case, perfect convergence is never guaranteed in finite-sample scenarios, so validation tools such as calibration and accuracy curves are indispensable in practical applications.

\textit{Amortized inference}. Following ideas from \textit{inference compilation} \cite{le2016inference} and \textit{pre-paid parameter estimation} \cite{mestdagh2019prepaid}, our method avoids fitting each candidate model to each data set separately. Instead, we cast the problem of model comparison as a supervised learning of absolute evidence and train a specialized neural network architecture to assign model evidences to each possible data set. This requires only the specification of plausible priors over each model's parameters and the corresponding forward process, from which simulations can be obtained on the fly. During the upfront training, we use online learning to avoid storage overhead due to large simulated grids or reference tables \cite{marin2018likelihood, mestdagh2019prepaid}. Importantly, the separation of model comparison into a costly upfront training phase and a cheap inference phase ensures that subsequent applications of the pre-trained networks to multiple observed data sets are very efficient. Indeed, we showed in our experiments that inference on thousands of data sets can take less than a second with our method. Moreover, by sharing and applying a pre-trained network for inference within a particular research domain, results will be highly reproducible, since the \textit{settings} of the method will be held constant in all applications. 

\textit{Raw data utilization and variable sample size}. The problem of insufficient summary statistics has plagued the field of approximate Bayesian computation for a long time, so as to deserve being dubbed the \textit{curse of insufficiency} \cite{marin2018likelihood}. Using sub-optimal summary statistics can severely compromise the quality of posterior approximations and validity of conclusions based on these approximations \cite{robert2011lack}. Our method avoids using hand-crafted summaries by aligning the architecture of the evidential neural network to the inherent probabilistic symmetry of the data \cite{bloem2019probabilistic}. Using specialized neural network architectures, such as permutation invariant networks or a combination of recurrent and convolutional networks, we also ensure that our method can deal with data sets containing variable numbers of observations. Moreover, by minimizing the strictly proper version of the logarithmic loss, we ensure that perfect convergence implies maximal data utilization by the network.

\textit{Absolute evidence and epistemic uncertainty}. Besides point estimates of model posterior probabilities, our evidential networks yield a full higher-order probability distribution over the posterior model probabilities themselves. By choosing a Dirichlet distribution, we can use the mean of the Dirichlet distribution as the best approximation of model posterior probabilities. Beyond that, following ideas from the study of subjective logic \cite{jsang2018subjective} and uncertainty quantification in classification tasks \cite{sensoy2018evidential}, we can extract a measure of \textit{epistemic uncertainty}. We employ epistemic uncertainty to quantify the impossibility of making a model selection decision based on a data set, which is classified as implausible under all candidate models. Therefore, the epistemic uncertainty serves as a proxy to measure absolute evidence, in contrast to relative evidence, as given by Bayes factors or posterior odds. This is an important practical advantage, as it allows us to conclude that all models in the candidate set are a poor approximation of the data-generating process of interest. Indeed, our initial experiments confirm that our measure of epistemic uncertainty increases when data sets no longer lie within the range of the considered models. However, extensive validation is needed in order to explore and understand which aspects of an observed sample lead to model misfit. Further, exploring connections to approaches using auxiliary probabilistic classifiers for detecting model misspecifications, such as the recent CARMEN method \cite{thomas2019diagnosing}, seems to be an interesting avenue for future research.

These advantageous properties notwithstanding, our proposed method has certain limitations. 
First, our regularized optimization criterion induces a trade-off between calibration and epistemic uncertainty, as confirmed by our experiments. 
This trade-off is due to the fact that we capture epistemic uncertainty via a special form of Kullback-Leibler (KL) regularization during the training phase, which renders the optimized loss function no longer strictly proper.
We leave it to future research to investigate whether this trade-off is fundamental and whether there are more elegant ways to quantify absolute evidence from a simulation-based perspective. 

Second, our method is intended for model comparison from a prior predictive (marginal likelihood) perspective. 
However, since we do not explicitly fit each model to data, we cannot perform model comparison/selection based on posterior predictive performance. 
We note that in certain scenarios, posterior predictive performance might be a preferable metric for model comparison, so in this case, simulation-based sampling methods should be employed (e.g., ABC or neural density estimation, \cite{da2018model, papamakarios2018sequential}). 

Third, perfect convergence might be hard to achieve in real-world applications. 
In this case, approximation error will propagate into model posterior estimates. 
Therefore, it is important to use performance diagnostic tools, such as calibration curves, accuracy of recovery, and overconfidence bounds, in order to detect potential estimation problems.
Finally, even though our method exhibits excellent performance on the domain examples considered in the current work, it might break down in high-dimensional parameter spaces. 
Future research should focus on applications to even more challenging model comparison scenarios, for instance, hierarchical Bayesian models with intractable likelihoods, or neural network models.

\section*{Acknowledgment}
This research was supported by the Deutsche Forschungsgemeinschaft (DFG, German Research Foundation; grant number GRK 2277 "Statistical Modeling in Psychology"). We thank the Technology Industries of Finland Centennial Foundation (grant 70007503; Artificial Intelligence for Research and Development) for partial support of this work. We also thank David Izydorczyk and Mattia Sensi for reading the paper and providing constructive feedback.

\bibliography{references}{}
\bibliographystyle{plain}

\clearpage

\section*{Appendix}
\setcounter{equation}{0}
\setcounter{figure}{0}
\setcounter{table}{0}
\setcounter{page}{1}
\setcounter{section}{1}
\makeatletter

\renewcommand{\thetable}{S\arabic{table}}
\renewcommand{\thefigure}{S\arabic{figure}}
\renewcommand{\thesubsection}{\Alph{subsection}}

\subsection{Neural Network Architectures}

As already discussed, we need specialized neural network architectures for dealing with data sets with variable numbers of observations $N$ and different probabilistic symmetries (e.g., \textit{i.i.d.} or temporal ordering). In the following, we describe the network architectures used to tackle the most common probabilistic symmetries observed in the social and life sciences, that is, exchangeable and temporal sequences.

\subsubsection*{Exchangeable Sequences}
\begin{figure*}[h]
\centering
\begin{subfigure}{.99\textwidth}
    \includegraphics[width=\textwidth]{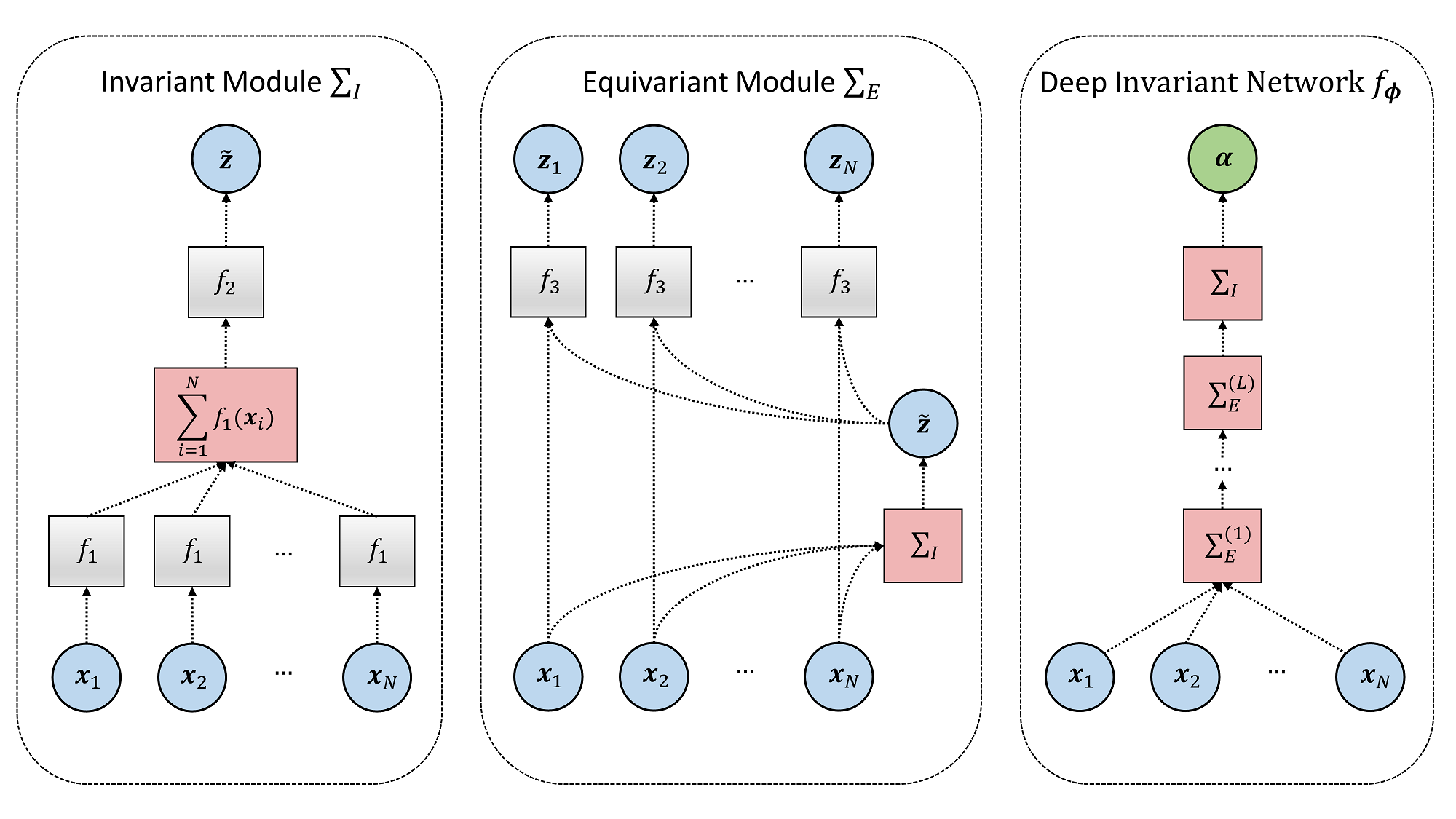}
\end{subfigure}
\caption[short]{The basic building blocks of a deep invariant evidential network encoding exchangeable sequences (adapted after \cite{bloem2019probabilistic}). The leftmost panel depicts an invariant module implementing an expressive permutation invariant function. The middle panel depicts an equivariant module implementing an expressive permutation equivariant function with a nested invariant module. The right panel depicts a deep invariant network consisting of a composition of equivariant modules followed by an invariant module with a ReLU output activation non-linearity.} \label{fig:Fig.2}
\end{figure*}

The most prominent feature of \textit{i.i.d.} sequences is \textit{permutation invariance}, since changing the order of individual elements does not change the associated likelihood function or posterior. Accordingly, if we denote an arbitrary permutation of $N$ elements as $\mathbb{S}_{N}(\cdot)$, the following should hold for the associated model posteriors:
\begin{align}
    p(\m \given \x_{1:N}) = p(\m \given \mathbb{S}_N(x_{1:N}))
\end{align}
We encode probabilistic permutation invariance by enforcing functional permutation invariance with respect to the outputs of the evidential network \cite{bloem2019probabilistic}. Following recent work on deep sets \cite{zaheer2017deep} and probabilistic symmetry \cite{bloem2019probabilistic}, we implement a deep invariant network performing a series of equivariant and invariant transformations. An invariant transformation is characterized by:
\begin{align}
    f(\mathbb{S}_N(\x_{1:N})) =  f(\x_{1:N})
\end{align}
that is, permuting the input elements does not change the resulting output. Such a transformation is often referred to as a pooling operation. On the other hand, an equivariant transformation is characterized by:
\begin{align}
    f(\mathbb{S}_N(\x_{1:N})) = \mathbb{S}_N(f(\x_{1:N}))
\end{align}
that is, permuting the input is equivalent to permuting the output of the transformation. We parameterize a learnable invariant function via an \textit{invariant module} performing a sequence of non-linear transformations followed by a pooling (sum) operation and another non-linear transformation:
\begin{align}
    \tilde{\bs{z}} = \Sigma_I(\x_{1:N}) = f_1\left( \sum_{i=1}^{N} f_2 (\x_i) \right)
\end{align}
where $f_1$ and $f_2$ can be arbitrary (non-linear) functions, which we parameterize via fully connected (FC) neural networks. \autoref{fig:Fig.2} (left panel) presents a graphical illustration of the invariant module. 

We parameterize a learnable equivariant transformation via an \textit{equivariant module} performing the following operations for each input element $i$: 
\begin{align}
    \bs{z}_i = \Sigma_E(\x_i, \tilde{\bs{z}}) = f_3(\x_i, \tilde{\bs{z}})
\end{align}
so that that $f_3$ is a combination of element-wise and invariant transforms (see \autoref{fig:Fig.2}, center). Again, we parameterize the internal function $f_3$ via a standard FC neural network. Note, that an equivariant module also takes as an input the output of an invariant module, in order to increase the expressiveness of the learned transformation. Thus, each equivariant module contains a separate invariant module. Finally, we can stack multiple equivariant modules followed by an invariant module, in order to obtain a deep invariant evidential network $f_{\bs{\phi}}: \mathcal{X}^N \rightarrow [1, \infty)^M$:
\begin{align}
    \bs{\alpha} &=  f_{\phib}(\x_{1:N}) = (\Sigma_I \circ \Sigma_E^{(L)} \circ \Sigma_E^{(L-1)} \circ \cdot\cdot\cdot \circ \Sigma_E^{(1)})(\x_{1:N})
\end{align}
where $\phib$ denotes the vector of all learnable neural network parameters and the final invariant module implements a shifted ReLU output non-linearity:
\begin{align}
\bs{\alpha} = \max \left(1, \Sigma_I(\bs{z}^{(L)}_{1:N}) \right)
\end{align}
in order to represent Dirichlet evidences ($0 < \alpha < 1$ is technically possible and valid but not of relevance for our purposes). The rightmost panel in \autoref{fig:Fig.2} provides a graphical illustration of a deep invariant network. We use this architecture in \textbf{Experiments 1, 2} and \textbf{4}.

\subsubsection*{Non-Exchangeable Sequences}

One of the most common non-exchangeable sequences encountered in practice are time-series with arbitrarily long temporal dependencies. A natural choice for time series-data with variable length are LSTM networks \cite{gers1999learning}, as recurrent networks are designed to deal with long sequences of variable size. Another reasonable choice might be 1D fully convolutional networks \cite{long2015fully}, which can also process sequences with variable length.  A different type of frequently encountered non-exchangeable data are images, which have successfully been tackled via 2D convolutional networks. 

\begin{figure*}
\centering
\begin{subfigure}{.99\textwidth}
    \includegraphics[width=\textwidth]{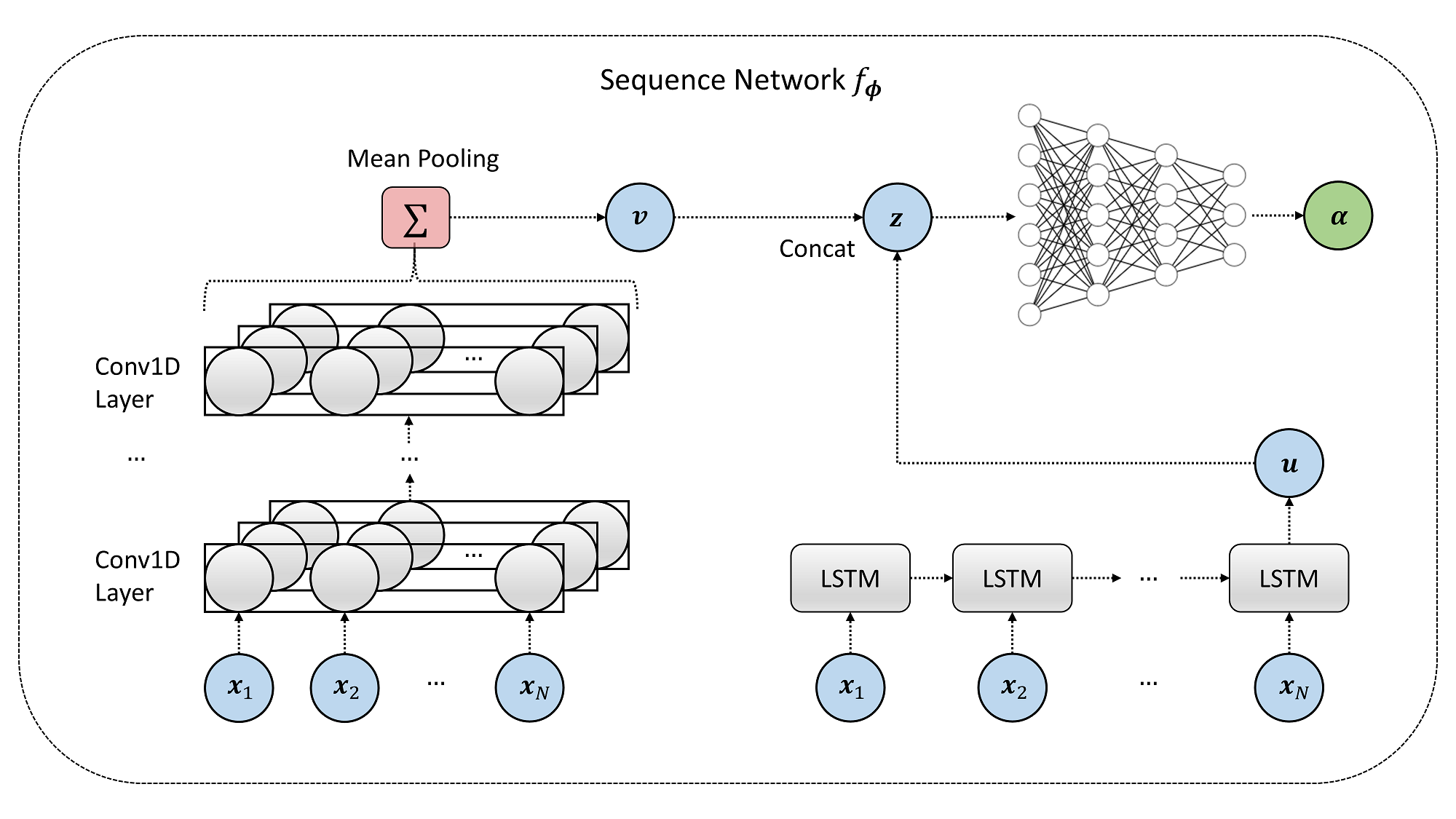}
\end{subfigure}
\caption[short]{The building blocks of an evidential sequence network. An input sequence generated from a model is fed thorough a number of convolutional layers in a hierarchical manner (left). The output of the final convolutional layer is passed through a mean pooling operator to obtain a fixed sized vector $\bs{v}$. The input sequence is also fed through an LSTM many-to-one recurrent neural network (right) to obtain another fixed-size vector $\bs{u}$. The vectors $\bs{v}$ and $\bs{u}$ are then concatenated into a vector $\bs{z}$ and passed through a standard fully connected network (upper right) to obtain the final Dirichlet evidences.} \label{fig:Fig.3}
\end{figure*}

Since, in this work, we apply our evidential method to time-series models, we will describe an architecture for processing data with temporal dependencies. 
We use a combination of a LSTM and a 1D convolutional network to reduce the observed time-series into fixed-size vector representations. 
We then concatenate these vectors and pass them through a standard fully connected network to obtain the final Dirichlet evidences. 
At a high level, our architecture performs the following operations on an input sequence $\x_{1:N}$. 
First, a many-to-one LSTM network reduces the input sequence to a vector $\bs{u}$ of pre-defined size. 
Then, a 1D convolutional network reduces the input sequence to a matrix $\bs{V} \in \mathbb{R}^{N' \textrm{x} W}$ where $N'$ is the length of the filtered sequence and $W$ is the number of filters in the final convolutional layer.
In this way, only the \textit{time} dimension of $V$ depends on the length of the input. A mean pooling operator is then applied to the time dimension of $\bs{V}$ to obtain a fixed-size representation $\bs{v}$ of size $W$.
Finally, the outputs of the LSTM and convolutional networks are concatenated and fed through a FC network $f$ with a shifted ReLU output non-linearity, which yields the Dirichlet evidences $\bs{\alpha}$. Thus, the computational flow is as follows:
\begin{align}
    \bs{u} &= \textrm{LSTM}(\x_{1:N}) \\
    \bs{v} &= \frac{1}{N'}\sum_{i=1}^{N'} \bs{V}_{i,:} \textrm{ with } \bs{V} = \textrm{Conv1D}(\x_{1:N}) \\
    \bs{z} &= \textrm{Concat}(\bs{u}, \bs{v}) \\
    \bs{\alpha} &= \max (1, f(\bs{z})) 
\end{align}
\autoref{fig:Fig.3} illustrates the computational flow of a sequence network. This architecture provides us with a powerful estimator for comparing models whose outputs consist of multivariate time-series. All neural network parameters $\bs{\phi}$ are jointly optimized during the training phase. We use this architecture in \textbf{Experiments 3} and \textbf{5}.

\subsection{Neural Network Training}

\noindent We train all neural networks described in this paper via mini-batch gradient descent\footnote{Code and simulation scripts for all experiments are available at \url{https://github.com/stefanradev93/BayesFlow}.}. 
For all following experiments, we use the Adam optimizer with a starter learning rate of $10^{-4}$ and an exponential decay rate of $.99$. 
We did not perform an extensive search for optimal values of network hyperparameters. 
All networks were implemented in Python using the \textit{TensorFlow} library \cite{abadi2016tensorflow} and trained on a single-GPU machine equipped with NVIDIA\textsuperscript{\textregistered} GTX1060 graphics card. See \textbf{Appendix A} for details on the neural network architectures.

\subsection{Performance Metrics}

\noindent Throughout the following examples, we use a set of performance metrics to assess the overall performance of our method. To test how well the method is able to recover the true model (hard assignment of model indices), we compute the accuracy of recovery as the fraction of correct model assignments over the total number of test datasets, where model assignments are done by selecting the model with the highest probability. 
To test how well the posterior probability estimates of the evidence network match the true model posteriors, we compute the expected calibration error (ECE, \cite{guo2017calibration}). 
The ECE measures the gap between the confidence and the accuracy of a classifier and is an unbiased estimate of exact miscalibration \cite{guo2017calibration}. 
In practice, we will report \textit{calibration curves} for each model, as these are easier to interpret for multi-class classification problems. 
Finally, to ensure that the method does not exhibit overconfidence, we compute an \textit{overconfidence metric} which is given by the difference between a high probability threshold $T$ (e.g., $T=0.95$) and the accuracy of the model above this threshold: $\textrm{overconfidence} =\max \{ 0, T - \frac{1}{ \given D_T \given }\sum_{i \in D_T}\mathds{1}_{[\widehat{m}^{(i)}=m^{(i)}]} \}$ where $D_T$ is the set of indices of predicted probabilities larger than $T$. 
Any deviation from zero would be indicative of overconfidence and thus lack of confidence in the method's estimates.

\subsection{Bonus Experiment: 400 Gaussian Mixture Models}

With this example, we want to show that our method is capable of performing model comparison on problems involving hundreds of competing models. Further, we want to corroborate the desired improvement in accuracy with increasing number of observations, as shown in the previous experiment.

\begin{figure*}
\centering
\begin{subfigure}{.49\textwidth}
    \includegraphics[width=\textwidth]{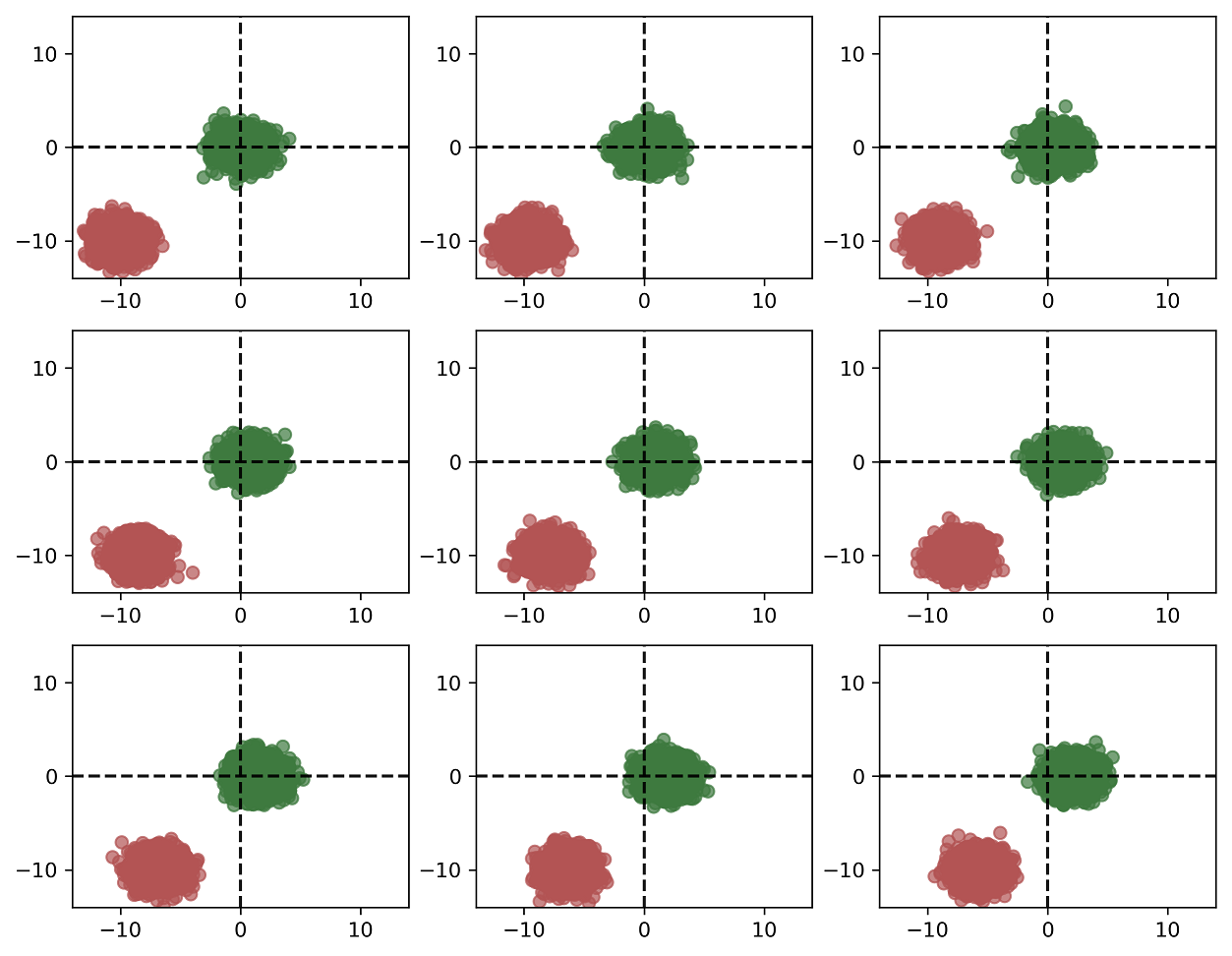}
    \caption{Example GMM datasets from 9 models}
    \label{fig:Fig.5a}
\end{subfigure}
\begin{subfigure}{.49\textwidth}
    \includegraphics[width=\textwidth]{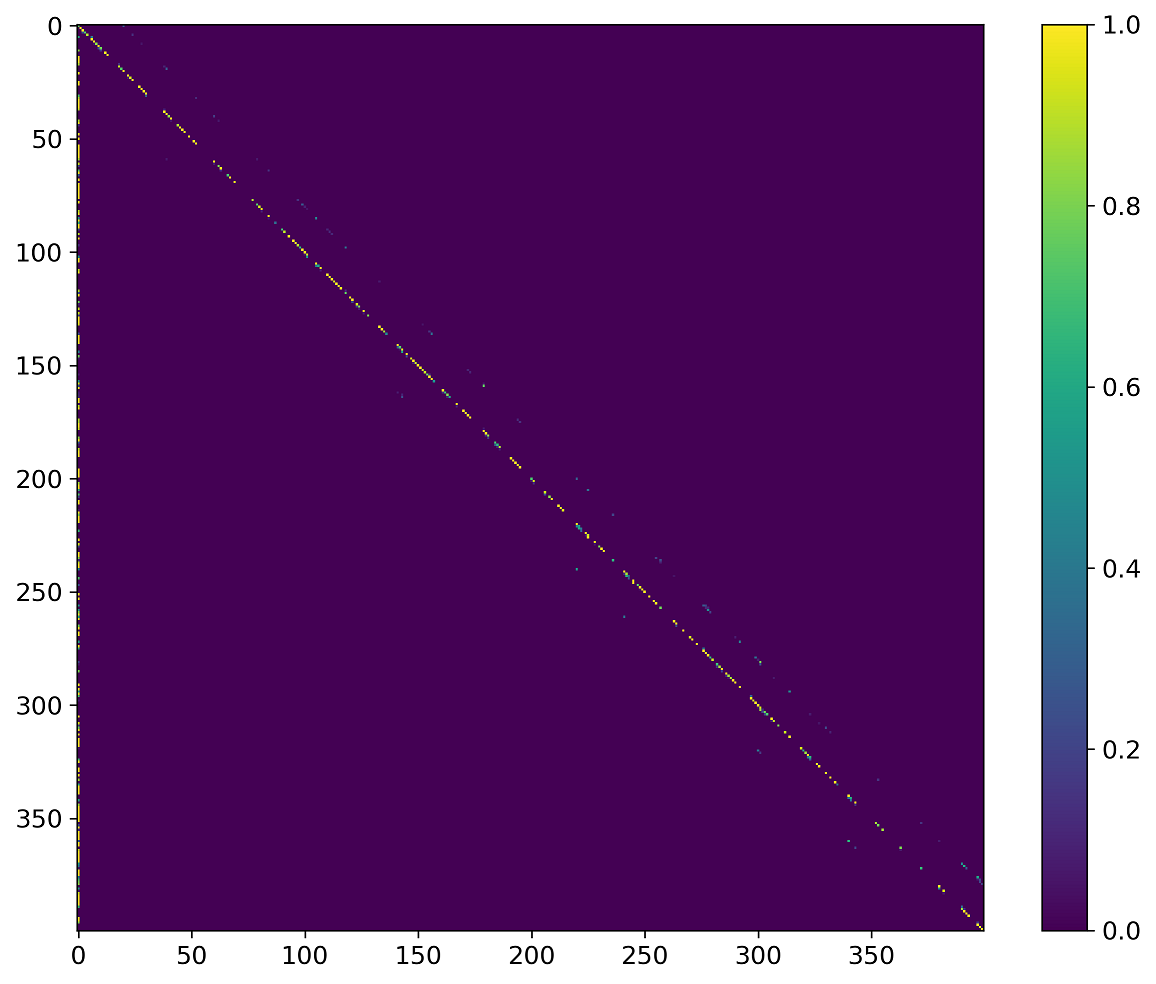}
    \caption{Confusion matrix at $N=250$}
    \label{fig:Fig.5b}
\end{subfigure}
\begin{subfigure}{.99\textwidth}
    \includegraphics[width=\textwidth]{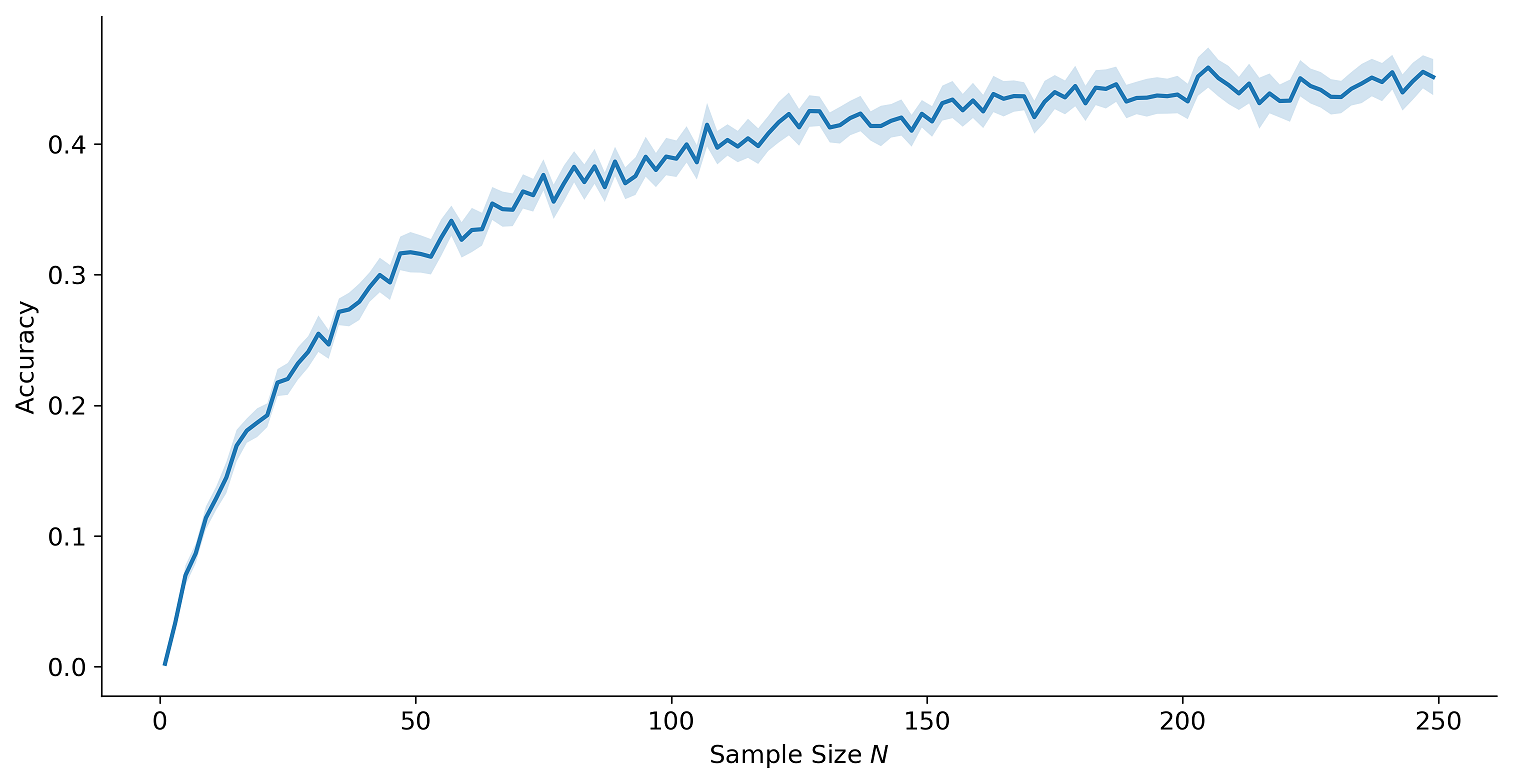}
    \caption{Accuracy over all $N$}
    \label{fig:Fig.5c}
\end{subfigure}
\caption[short]{Data and results from \textbf{Experiment 2}. \textbf{(a)} 9 example data-sets from 9 different GMM models. The X-coordinates of both clusters increase linearly left to right, and coordinate differences are very small; \textbf{(b)} Heatmap of the confusion matrix between true and predicted model indices. Accuracy of recovery is color-coded according to the colorbar depicted  at the right. $N=250$; \textbf{(c)} Accuracy of recovery as a function of sample size $N$.} \label{fig:Fig.5}
\end{figure*}

To this aim, we construct a setting with $400$ 2D Gaussian Mixture Models (GMMs) with $2$ mixture components. The construction of the models and data proceeds as follows. We first specify the $400$ mean vectors on two linearly-spaced $20 \times 20$ grids: $\bs{\mu}_0^{(m)} \in [-10, 0] \times [-10, 0]$ and $\bs{\mu}_1^{(m)} \in [0, 10] \times [0, 10]$ for $m = 1,\dots,400$. We then generate data from each GMM model by:
\begin{align}
N &\sim \mathcal{U}_D(1, 250) \\
\pi^{(m)} &\sim \textrm{Beta}(30, 30) \\
k &\sim \textrm{Bernoulli}(\pi^{(m)}) \\
\x_j^{(m)} &\sim \mathcal{N}(\bs{\mu}_k^{(m)}, \mathbb{I}) \textrm{ for } j = 1,\dots,N
\end{align}
where $\mathcal{U}_D$ denotes the discrete uniform distribution and $\mathbb{I}$ the identity matrix of appropriate dimension. Thus, each dataset consists of $N$ $i.i.d.$ samples from one of $400$ GMMs with different component mean vectors. \autoref{fig:Fig.5a} depicts simulated datasets from $9$ GMM with linearly increasing $X$-coordinates showing that differences between the models are very subtle.

We train an invariant evidential network for $30$ epochs for approximately $1.2$ hours wall-clock time. We then validate the performance of the network on all possible $N$ between $1$ and $250$ with $5000$ previously unseen simulated datasets. \autoref{fig:Fig.5b} depicts a heatmap of the normalized \textit{confusion matrix} obtained on the validation data sets with $N=250$. Importantly, we observe that the main diagonal of the heatmap indicates that the predicted model indices mostly match the true model indices, with mistakes occurring mainly between models with very close means. Bootstrap mean accuracy at $N=250$ was around $0.451$ (\textit{SD}=0.007), which is $180$ times better than chance accuracy ($0.0025)$ considering the dimensionality of the problem. Finally, \autoref{fig:Fig.5c} depicts the accuracy of recovery over all $N$ considered. Again, we observe that accuracy improves as more observations become available, with sub-linear scaling.

\subsection{Priors for Experiment 4}

The prior parameter distributions for all models used in Experiment 4 are listed in \autoref{table:Table 1}.

\begin{table*}[h]
\centering
\caption{Parameter priors of the free parameters and numerical values of the fixed parameters of the six EAM models considered in \textbf{Experiment 4}.}
\begin{tabular}{llllllllll}
\toprule
  & $v_1$ & $v_2$ & $a$  & $t_0$ & $z_r$ & $\alpha$ & $s_{t_0}$ & $s_{v}$ & $s_{zr}$ \\
\midrule
$\mathcal{M}_1$ & $\mathcal{U}(0, 6)$ & $\mathcal{U}(-6, 0)$  & $\mathcal{U}(0.6, 3)$ &  $\mathcal{U}(0.2, 1.5)$ & $0.5$ & 0 & 0 & 0 & 0\\
$\mathcal{M}_2$ & $\mathcal{U}(0, 6)$ & $\mathcal{U}(-6, 0)$  & $\mathcal{U}(0.6, 3)$ & $\mathcal{U}(0.2, 1.5)$ & $\mathcal{U}(0.3, 0.7)$ &  0 & 0 & 0 & 0\\
$\mathcal{M}_3$ & $\mathcal{U}(0, 6)$ & $\mathcal{U}(-6, 0)$  & $\mathcal{U}(0.6, 3)$ & $\mathcal{U}(0.2, 1.5)$ & $\mathcal{U}(0.3, 0.7)$  & $\mathcal{U}(1, 2)$ & 0 & 0 & 0\\
$\mathcal{M}_4$ & $\mathcal{U}(0, 6)$ & $\mathcal{U}(-6, 0)$  & $\mathcal{U}(0.6, 3)$ &  $\mathcal{U}(0.2, 1.5)$ & $\mathcal{U}(0.3, 0.7)$  & $\mathcal{U}(1, 2)$ & $\mathcal{U}(0, 0.4)$ & 0 & 0\\
$\mathcal{M}_5$ & $\mathcal{U}(0, 6)$ & $\mathcal{U}(-6, 0)$  & $\mathcal{U}(0.6, 3)$ &  $\mathcal{U}(0.2, 1.5)$ & $\mathcal{U}(0.3, 0.7)$  & $\mathcal{U}(1, 2)$ & $\mathcal{U}(0, 0.4)$ & $\mathcal{U}(0, 2)$ & 0\\
$\mathcal{M}_6$ & $\mathcal{U}(0, 6)$ & $\mathcal{U}(-6, 0)$  & $\mathcal{U}(0.6, 3)$ &  $\mathcal{U}(0.2, 1.5)$ & $\mathcal{U}(0.3, 0.7)$  & $\mathcal{U}(1, 2)$ & $\mathcal{U}(0, 0.4)$ & $\mathcal{U}(0, 2)$  & $\mathcal{U}(0, 0.6)$ \\
\bottomrule
\end{tabular}
\label{table:Table 1}
\end{table*}

\subsection{Details for Experiment 4}

The forward model is formulated as a set of five ordinary differential equations (ODEs) describing how the neuron membrane potential $V(t)$ unfolds in time as a function of an injected current $I_{inj}(t)$, and ion channels properties.
The change in membrane potential is defined by the membrane ODE: 
\begin{align}
    C \frac{dV}{dt} &= -I_L - I_{Na} - I_K - I_M + I_{inj} + \sigma \eta (t)
\end{align}
where $C$ is the specific membrane capacitance, $\sigma \eta (t)$ the intrinsic neural noise, and the $I_j$s are the ionic currents
flowing through channels, such that:

\begin{align}
    I_L &= g_L(V - E_L) \\
    I_{Na} &= \Bar{g}_{Na} m^3 h (V - E_{Na}) \\
    I_K &= \Bar{g}_K n^4 (V - E_K) \\
    I_M &= \Bar{g}_M p (V - E_M).
\end{align}
Here, $g_L$ is the leak conductance, while $\Bar{g}_{Na},\Bar{g}_K,\Bar{g}_M$ are the sodium, potassium and
M-type channel maximum conductances, respectively. $E_L,E_{Na}$ and $E_K$ denote the leak equilibrium potential, the sodium
and potassium reversal potentials, respectively. In particular, $g_L$ is assumed constant through time, whilst the other conductances vary over time. Consistently, $(m,h,n,p)$ indicates the vector of the state variables accounting for ion channel gating kinetics evolving according to the following set of ODEs:

\begin{align}
    \frac{di}{dt} &= \alpha_i(V)(1-i) - \beta_i(V)i \\
    \frac{dp}{dt} &= \frac{p_{\infty}(V)-p}{\tau_p(V)} 
\end{align}
where $i\in \{m,h,n\}$, and $\alpha_i(V)$, $\beta_i(V)$, $p_{\infty}(V)$ and $\tau_p(V)$ are nonlinear functions of the membrane potential (see \cite{pospischil2008minimal} for details).

\begin{figure*}
\centering
\begin{subfigure}{.99\textwidth}
    \includegraphics[width=\textwidth]{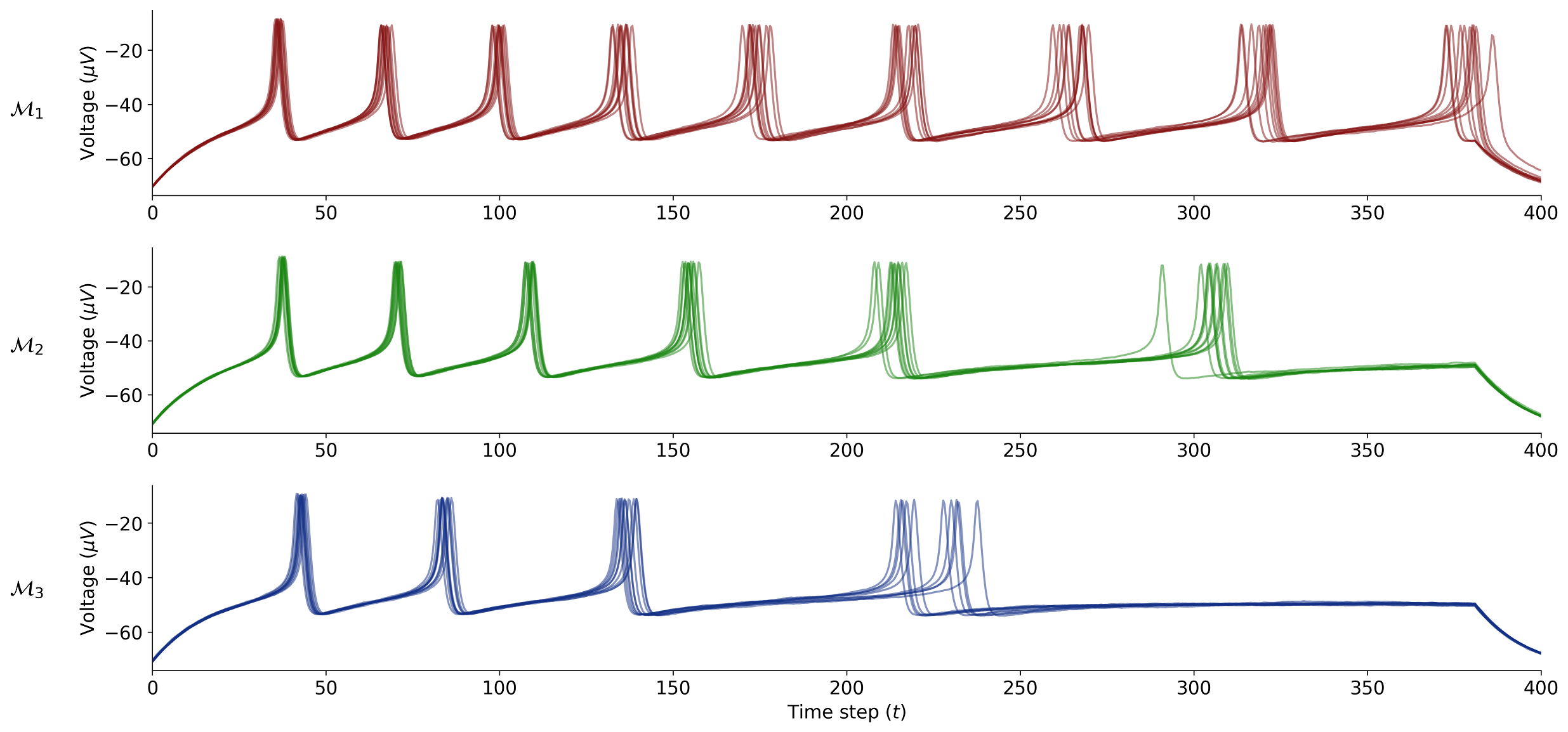}
    \caption{Example spiking patterns from all three models}
    \label{fig:Fig.8a}
\end{subfigure}
\begin{subfigure}{.33\textwidth}
    \includegraphics[width=\textwidth]{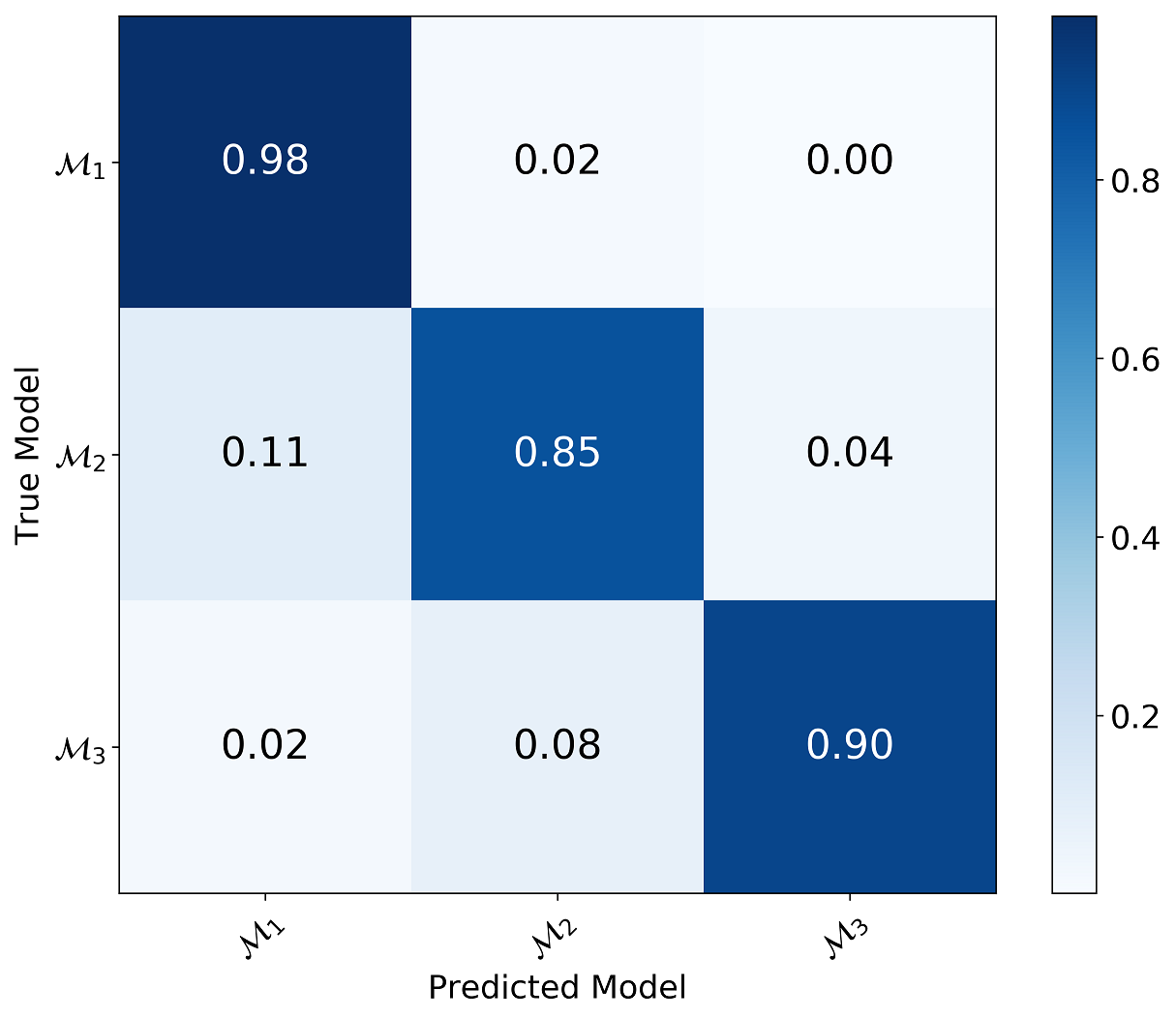}
    \caption{Confusion matrix with $T=100$}
    \label{fig:Fig.8b}
\end{subfigure}
\begin{subfigure}{.62\textwidth}
    \includegraphics[width=\textwidth]{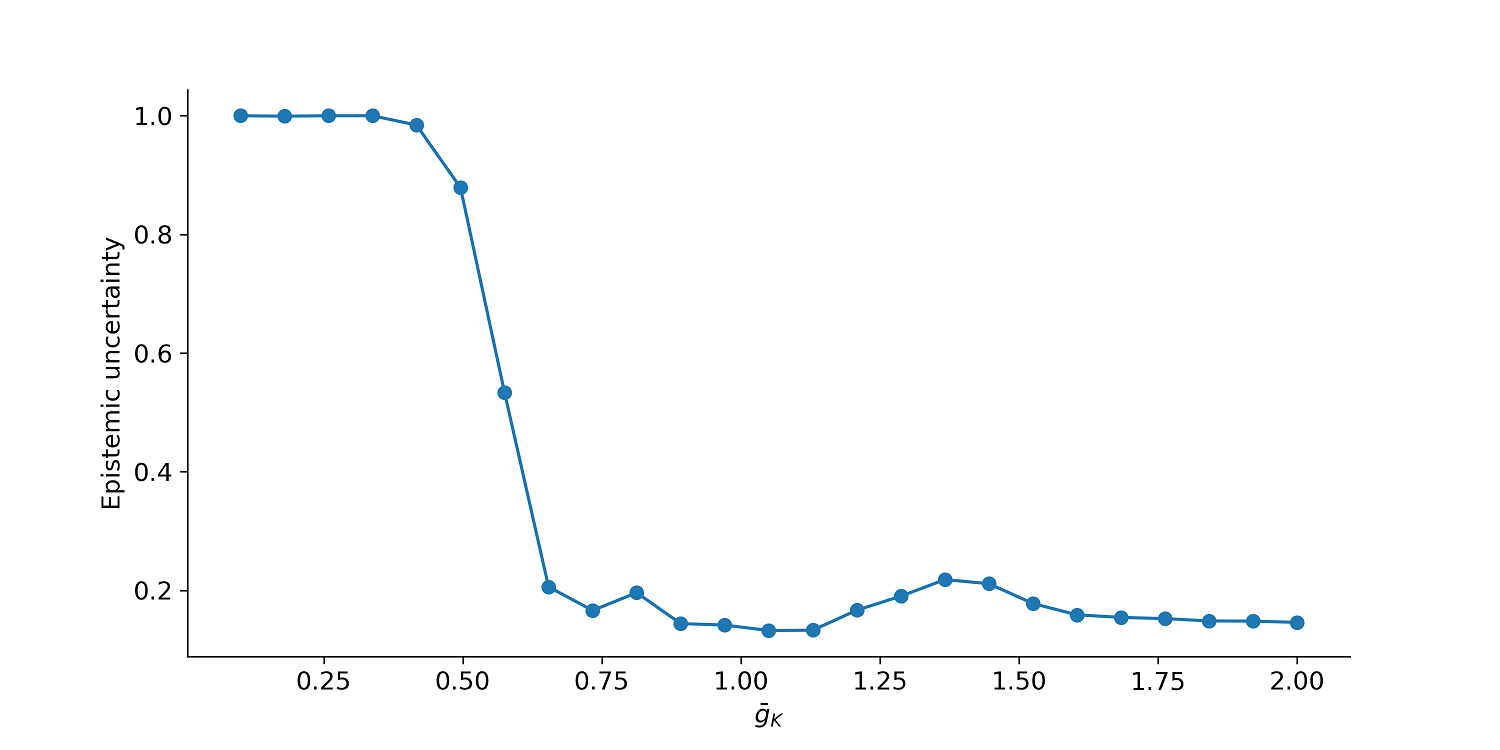}
    \caption{Epistemic uncertainty as a function of $\bar{g}_K$}
    \label{fig:Fig.8c}
\end{subfigure}
\begin{subfigure}{.99\textwidth}
    \includegraphics[width=\textwidth]{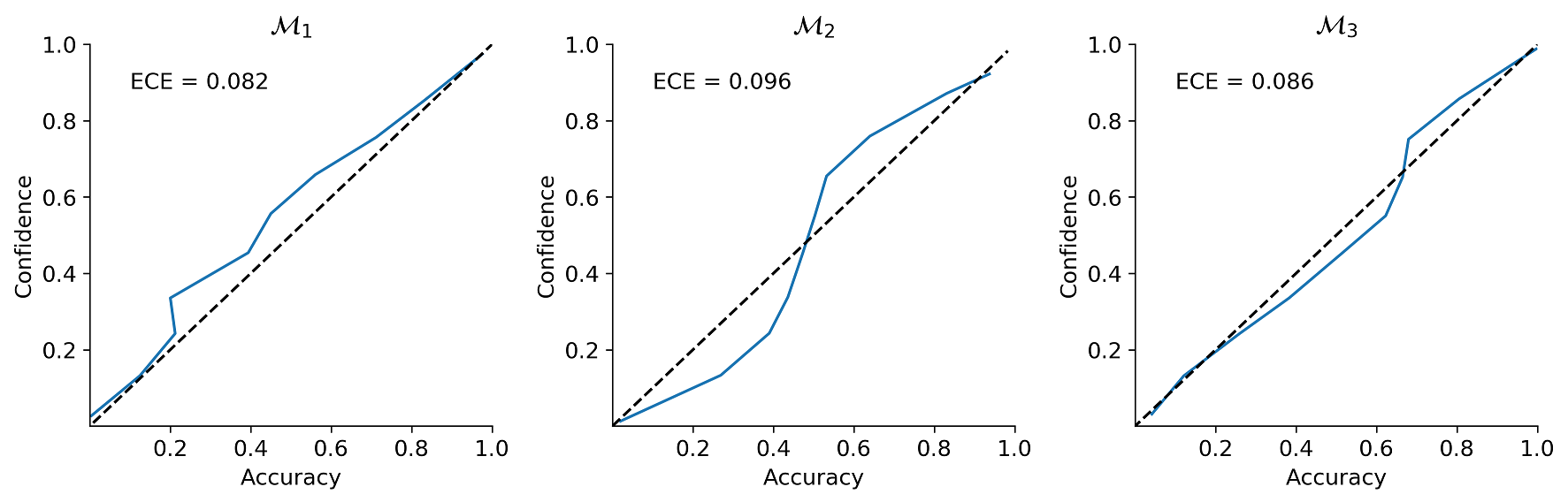}
    \caption{Calibration at $T=400$}
    \label{fig:Fig.8d}
\end{subfigure}
\caption{Example Hodgkin-Huxley spiking patterns and validation results from \textbf{Experiment 5}.}\label{fig:Fig.8}
\end{figure*}

In our simulated experiment, we treat conductances $g_L,\Bar{g}_{Na},\Bar{g}_K$ and $\Bar{g}_M$ as free parameters, and consider different neuronal models based on different parameter configurations. It is also assumed that such configurations allow to affect the span of the possible firing patterns attainable by each model. In particular, we consider 3 models $\bs{\mathcal{M}} = \{\mathcal{M}_1, \mathcal{M}_2, \mathcal{M}_3\}$ defined by the parameter sets $\thetab_1 = (\Bar{g}_{Na},\Bar{g}_K)$,  $\thetab_2 = (\Bar{g}_{Na},\Bar{g}_K,\Bar{g}_M)$ and $\thetab_1 = (\Bar{g}_{Na},\Bar{g}_K,\Bar{g}_M,g_L)$. We place the following priors over the parameters $\Bar{g}_{Na}$ and $\Bar{g}_K$:
\begin{align}
    \Bar{g}_{Na} &\sim \mathcal{U}(1.5, 30) \\
    \Bar{g}_K &\sim \mathcal{U}(0.3, 15)
\end{align}
When not considered as free parameters, $\Bar{g}_M$ and $g_L$ are set to default values, such that $\Bar{g}_M=0.07$ and $g_L=0.1$, otherwise parameters are drawn from the following priors:
\begin{align}
    \Bar{g}_M &\sim \mathcal{U}(0.005, 0.3) \\
    \Bar{g}_L &\sim \mathcal{U}(0.01, 0.18)
\end{align}
\autoref{fig:Fig.8a} depicts $50$ example runs from each model with respective parameters $\thetab_1 = (3.0, 2.0)$, $\thetab_2 = (3.0, 2.0, 0.1)$, $\thetab_3 = (3.0, 2.0, 0.1, 0.11)$.

\end{document}